\documentclass[conference]{IEEEtran}
\usepackage{amsmath,amsfonts}
\usepackage{algorithmic}
\usepackage{algorithm}
\usepackage{array}
\usepackage{longtable}
\usepackage[caption=false,font=normalsize,labelfont=sf,textfont=sf]{subfig}
\usepackage{textcomp}
\usepackage{stfloats}
\usepackage{url}
\usepackage{verbatim}
\usepackage{graphicx}
\usepackage{cite}
\usepackage{soul}
\usepackage{multirow}
\usepackage{xcolor}
\usepackage{svg}
\usepackage{hyperref}
\usepackage{lipsum}
\usepackage{pifont}
\usepackage{makecell}
\usepackage{booktabs}

\renewcommand{\arraystretch}{1.2}

\makeatletter
\newcommand{\linebreakand}{%
  \end{@IEEEauthorhalign}
  \hfill\mbox{}\par
  \mbox{}\hspace{90pt}\begin{@IEEEauthorhalign}
}
\makeatother

\renewcommand{\textcolor}[2]{#2}  % disables all colors globally

\begin{document}

\title{Brain Inspired Probabilistic Occupancy Grid Mapping with Vector Symbolic Architectures}

\author{\IEEEauthorblockN{Shay Snyder}
\IEEEauthorblockA{\textit{George Mason University}\\
Fairfax, USA\\
ssnyde9@gmu.edu}
\and
\IEEEauthorblockN{Andrew Capodieci}
\IEEEauthorblockA{\textit{Neya Systems, LLC}\\
Warrendale, USA \\
acapodieci@neyarobotics.com}
\linebreakand
\IEEEauthorblockN{David Gorsich}
\IEEEauthorblockA{\textit{US Army Futures Command}\\
Warren, USA \\
david.j.gorsich.civ@army.mil}
\and
\IEEEauthorblockN{Maryam Parsa}
\IEEEauthorblockA{\textit{George Mason University}\\
Fairfax, USA \\
mparsa@gmu.edu}
}

% \author{Shay Snyder, Andrew Capodieci, David Gorsich, and Maryam Parsa
% % <-this % stops a space
% % \thanks{Manuscript received April 19, 2021; revised August 16, 2021.}

% }

% \markboth{}%
% {Shell \MakeLowercase{\textit{Snyder et al.}}: Brain Inspired Probabilistic Occupancy Grid Mapping with Hyperdimensional Computing}

%\IEEEpubid{0000--0000/00\$00.00~\copyright~2024 IEEE}

\maketitle
\thispagestyle{plain}
\pagestyle{plain}

\begin{abstract}
Real-time robotic systems require advanced perception, computation, and action capability.
However, the main bottleneck in current autonomous systems is the trade-off between computational capability, energy efficiency and model determinism.
World modeling, a key objective of many robotic systems, commonly uses occupancy grid mapping (OGM) as the first step towards building an end-to-end robotic system with perception, planning, autonomous maneuvering, and decision making capabilities.
OGM divides the environment into discrete cells and assigns probability values to attributes such as occupancy and traversability. Existing methods fall into two categories: traditional methods and neural methods.
Traditional methods rely on dense statistical calculations, while neural methods employ deep learning for probabilistic information processing.
In this study, we propose a vector symbolic architecture-based OGM system (VSA-OGM) that retains the interpretability and stability of traditional methods with the improved computational efficiency of neural methods.
Our approach, validated across multiple datasets, achieves similar accuracy to covariant traditional methods while reducing latency by \textcolor{blue}{approximately} \textcolor{blue}{45x} and memory by \textcolor{blue}{400x}. Compared to invariant traditional methods, we see similar accuracy values while reducing latency by \textcolor{blue}{5.5x}. \textcolor{blue}{Moreover, we achieve up to 6x latency reductions compared to neural methods while eliminating the need for domain-specific model training.}
\textcolor{blue}{This work demonstrates the potential of vector symbolic architectures as a practical foundation for real-time probabilistic mapping in autonomous systems operating under strict computational and latency constraints.}
\end{abstract}


% Temporarily suppress footnote number
\renewcommand{\thefootnote}{}
\footnotetext{DISTRIBUTION A. Approved for public release; distribution unlimited. OPSEC \#8540.}
\addtocounter{footnote}{-1} % So it doesn't increment the counter
\renewcommand{\thefootnote}{\arabic{footnote}} % Restore if needed

\section{Introduction}

\noindent
An essential capability for real-time robotic systems is the accurate and computationally efficient probabilistic mapping of environments.
Efficient probabilistic mapping is the central idea within OGM where environments are discretized into individual cells, \textcolor{blue}{also known as pixels and voxels in 2D and 3D environments, respectively~\cite{10273187}.}
Each cell is then assigned labels or properties such as occupancy, traversability, or other relevant attributes based on \textcolor{blue}{application-specific} requirements~\cite{wilson2022convolutional}.
OGM methods are broadly categorized into two main areas: traditional and neural methods.
Traditional methods~\cite{thrun2002probabilistic, zhi2019continuous, elfes1989using, senanayake2017bayesian, Senanayake_Ramos_2018, duong2022autonomoussbkm, Ghaffari_Jadidi2018-vg, doherty2016probabilistic, 11099540} adopt a dense approach where each \textcolor{blue}{cell} is assigned a fully defined probability distribution or feature representation.

\textcolor{blue}{In contrast, neural methods, such as~\cite{evilog, 9879943, 8953655, xiang2017darnn, wu2020motionnet}, utilize deep learning architectures, such as convolutional neural networks~\cite{evilog, devilog}, neural radiance fields~\cite{carlson2023cloner}, and large language models~\cite{Yang_Liu_Zhang_Pan_Guo_Li_Chen_Gao_Li_Guo_Zhang_2025}, to create a black-box solution.}
\textcolor{blue}{This approach mitigates many of the computational challenges associated with traditional methods but introduces fragility to domain adaptation and model stochasticity through domain-specific pretraining with backpropagation~\cite{10017290}.}

Although neural methods present computational advantages over traditional approaches, their intrinsic stochasticity and lack of interpretability pose significant validation challenges, particularly in safety-critical applications~\cite{8342158}. Moreover, recent neural \textcolor{blue}{methods} leverage foundation models to perform probabilistic inference, making them unsuitable for real-time deployment on edge devices with computational and power constraints~\cite{Yang_Liu_Zhang_Pan_Guo_Li_Chen_Gao_Li_Guo_Zhang_2025}.

Recent advances in cognitive science~\cite{plate2003holographic, neuro-hdc} and computational neuroscience~\cite{eliasmith2013build, dumont2025symbols} highlight the unique computational properties \textcolor{blue}{of} hyperdimensional spaces.
Also known as vector symbolic architectures (VSA), these methods are actively being explored as mathematical analogs to the higher-order computational capabilities of spiking neurons~\cite{eliasmith2003neural, dumont2025symbols}.
Furthermore, VSA approaches posses unique computational properties allowing them to statistically, and deterministically, encode semantic and spatial information~\cite{KomerBrent2020, gobin2024exploration, dumont2025symbols}.

In this work, we introduce a VSA framework, VSA-OGM, leveraging the unique probabilistic properties of bio-inspired hyperdimensional vectors to efficiently and accurately perform OGM. Our approach draws inspiration from both \textcolor{blue}{traditional}~\cite{thrun2002probabilistic} and neural~\cite{evilog} methods, aiming to strike a balance between the accuracy, interpretability, stability, and domain flexibility of traditional methods with the computational efficiency of neural methods. We refer to this intersection of traditional and neural methods as neuro-symbolic methods.

To the best of our knowledge, VSA-OGM is the first large scale application and development of a VSA framework for real-time OGM. Throughout the remainder of this work, we provide further background information on the fields of OGM and VSAs. We discuss the specifics of VSA-OGM and highlight the performance of our framework on three synthetic datasets~\cite{senanayake2017bayesian, evilog} and one real dataset~\cite{Radish} compared to \textcolor{blue}{multiple} traditional \textcolor{blue}{methods}~\cite{senanayake2017bayesian, zhi2019continuous, duong2022autonomoussbkm, Ghaffari_Jadidi2018-vg, doherty2016probabilistic}, and one neural method~\cite{evilog}. We also conduct an in-depth ablation study providing further insight into the capabilities and unique aspects of VSA-OGM.
We conclude with a detailed discussion of our results and potential avenues for future research. In summary, the major contributions of our paper are:

\begin{itemize}
    \item We introduce a \textcolor{blue}{neuro-symbolic} VSA framework for OGM, VSA-OGM, that maintains the accuracy of traditional methods while leveraging the improved computational characteristics of neural methods.
    \item Compared to traditional covariant methods that explicitly maintain covariance between \textcolor{blue}{cells}~\cite{senanayake2017bayesian, Ghaffari_Jadidi2018-vg}, across two simulated~\cite{senanayake2017bayesian} and one real-world~\cite{Radish} dataset, \textcolor{blue}{VSA-OGM} achieves approximately 400x and 45x reductions in memory usage and latency, respectively, while maintaining comparable accuracy.
    \item We demonstrate the single- and multi-agent mapping capabilities of VSA-OGM compared to two invariant traditional methods~\cite{zhi2019continuous, duong2022autonomoussbkm}, which do not maintain covariance between \textcolor{blue}{cells}, achieving approximately 5.5x lower latency while preserving comparable accuracy.
    \item We highlight the capabilities of \textcolor{blue}{VSA-OGM}, which achieves approximately \textcolor{blue}{6x} lower latency than neural methods~\cite{evilog}, without requiring model pretraining and maintaining comparable \textcolor{blue}{mapping performance}. 
    \item \textcolor{blue}{We conduct a comprehensive ablation study to evaluate the performance impact of various hyperparameter and framework configurations in VSA-OGM. Our results clarify the role of each component and reveal how their integration forms a neurosymbolic framework for OGM. This enables VSA-OGM to retain the accuracy and determinism of traditional methods while surpassing the computational efficiency of neural methods, without requiring extensive pretraining or introducing domain fragility.}
\end{itemize}
\section{Background \& Motivation}

% Summarize breaking the section into two main portions
\noindent
%We divide the background into two subsections. 
We begin this section with a discussion on the details of occupancy grid mapping (OGM) and the breakdown \textcolor{blue}{of OGM implementations} into traditional, neural, and \textcolor{blue}{neuro-symbolic} methods. We continue with a background of vector symbolic architectures (VSAs), their application in cognitive neuroscience, and on-going research initiatives. Moreover, we discuss how VSAs could serve as a brain-inspired, \textcolor{blue}{neuro-symbolic framework} providing high-levels of accuracy from traditional methods and computational efficiency from neural methods.

\subsection{Occupancy Grid Mapping (OGM)}

% ------------------------------------
% Discuss the basic properties of OGM
% ------------------------------------
OGM discretizes continuous, n-dimensional environments into discrete \textcolor{blue}{cells}, commonly known as \textcolor{blue}{pixels or voxels in 2D and 3D environments, respectively}~\cite{wilson2022convolutional}. \textcolor{blue}{Cells} are assigned arrays of attributes, ranging from \textcolor{blue}{binary} indicators of occupancy to more complex attributes, such as cost values representing the likelihood of traversal through a region.
\textcolor{blue}{OGM has diversified into several distinct approaches, with each having its own operational characteristics.
\textit{Traditional methods}, such as those described in~\cite{thrun2002probabilistic, elfes1989using}, represent the environment using dense grids of statistical distributions, where each cell explicitly models occupancy probabilities. More recent traditional methods, such as~\cite{senanayake2017bayesian, Senanayake_Ramos_2018, zhi2019continuous}, project probabilistic information into a continuous Hilbert space to improve computational and space complexity. In contrast, \textit{neural methods}~\cite{evilog, devilog, schreiber2022multi, lee2024inverse} leverage deep learning to encode environmental information into network weights, where probabilistic relationships are captured implicitly through hidden representations.}

\textcolor{blue}{In this work, we introduce a novel class of OGM algorithms, which we term neuro-symbolic methods. These approaches seek to combine the efficiency of traditional methods with the representational power of neural methods, while minimizing stochasticity and eliminating the need for domain-specific pretraining. Such characteristics make them particularly well-suited for mission-critical and safety-sensitive applications. Within this emerging class, we present VSA-OGM, a framework built on brain-inspired hyperdimensional vectors and deterministic operations that project probabilistic inference into high-dimensional space. By bridging the gap between traditional and neural methods, VSA-OGM establishes a foundation for neurosymbolic OGM and represents a novel contribution to the family of OGM algorithms. We explore each OGM family in detail in the following sections, with a high-level comparison summarized in Table~\ref{tab:ogm-area}.}

{
\renewcommand{\arraystretch}{1.2}
\begin{table}[]
    \centering
    \caption{Comparing the determinism, accuracy, runtime complexity, and space complexity of traditional, neural, and \textcolor{blue}{neuro-symbolic} occupancy grid mapping approaches}

    \begin{tabular}{c|ccc}
    % \cline{1-4}
    \hline
    \textbf{Method}                                                & \textbf{Traditional} & \textbf{Neural} & \textbf{Neuro-Symbolic} \\ \hline
    \textbf{Determinism}                                           & Very High            & Low             & High                    \\ 
    \begin{tabular}[c]{@{}l@{}}\textbf{Accuracy}\end{tabular}      & Very High            & High            & High                    \\
    \begin{tabular}[c]{@{}l@{}}\textbf{Runtime Comp.}\end{tabular} & $\mathrm{O}(n^3)$    & Varies          & Constant                \\ 
    \begin{tabular}[c]{@{}l@{}}\textbf{Space Comp.}\end{tabular}   & $\mathrm{O}(n^2)$    & Varies          & Constant                \\ \hline
    % \hline
    \end{tabular}

    \label{tab:ogm-area}
\end{table}
}

\subsubsection{Traditional Methods}

\textcolor{blue}{Traditional OGM methods, introduced by Elfes~\cite{elfes1989using}, rely on a dense, grid-based discretizations of the environment, where each cell stores a probabilistic estimate of occupancy. These methods use geometric sensor models with Bayesian updates to recursively estimate occupancy from sensor data, providing a deterministic and interpretable map representation. However, the grid's fixed resolution can limit accuracy and scalability, especially in large or complex environments, since computational and memory complexity scales linearly with the number of cells.}

\textcolor{blue}{To overcome the resolution limitations of discrete grids, continuous OGM methods have been developed, which approximate occupancy as a spatially continuous function rather than discrete cells. Among these, Gaussian Process (GP)-based OGM is a well-established approach that models spatial correlations and uncertainty directly from sensor data~\cite{Ghaffari_Jadidi2018-vg}. While GPs provide smooth and high-fidelity occupancy estimates, their computational complexity grows cubically with the number of observations, versus the size of the environment, making them challenging to apply directly in real-time settings with high-fidelity sensors (i.e. high density point clouds)~\cite{Ghaffari_Jadidi2018-vg}.}

\textcolor{blue}{Bayesian Hilbert Maps (BHMs)\cite{senanayake2017bayesian} were introduced as an alternative to GP-based OGM approches that represent occupancy as projections onto a sparse set of basis functions in Hilbert space. BHMs avoid storing explicit observations and reduce runtime and memory costs significantly while maintaining probabilistic interpretability. Fast-BHM\cite{zhi2019continuous} further improves computational performance, to sub-quadratic levels, by assuming independence between basis functions to further approximate the Bayesian update.}

\textcolor{blue}{More recent works such as, Sparse Bayesian Kernel-based OGM (SBKM)~\cite{duong2022autonomoussbkm} extend this line of work by focusing on sparse structural features within the environment, such as lines and curves, rather than dense occupancy estimates. This makes SBKM especially useful for efficiently capturing high-level geometry relevant to tasks like path planning and map simplification, though it is not intended as a direct replacement for dense occupancy grids~\cite{duong2022autonomoussbkm}.}

% -------------------------------------------
% Discuss the fundamentals of neural methods
% -------------------------------------------
\subsubsection{Neural Methods}

\textcolor{blue}{Neural methods, also referred to as \textit{neural implicit representations~\cite{zhang2024rapid}}, utilize deep learning to encode spatial information within the weights of neural networks~\cite{evilog}. These methods map sensory inputs into continuous, high-dimensional latent spaces, contrasting with traditional methods that explicitly store occupancy or semantic attributes at discrete locations. Several distinct network architectures have been proposed:}
\begin{itemize}
    \item \textcolor{blue}{Neural Radiance Fields (NeRF)~\cite{9879943} employ multi-layer perceptrons (MLPs) with positional encodings to map 3D coordinates and viewing directions into volumetric color and density estimates. This enables continuous and differentiable scene representations, primarily for view synthesis and rendering tasks.}
    \item \textcolor{blue}{Occupancy Networks~\cite{8953655} similarly rely on MLPs but focus on learning implicit occupancy functions. Given a 3D point, the network outputs a probability of occupancy, effectively defining decision boundaries that describe object geometry at arbitrary resolutions.}
    \item \textcolor{blue}{Convolutional Neural Networks (CNNs)~\cite{wu2020motionnet} are typically used for grid-based inputs such as LIDAR or camera data. These networks exploit local spatial correlations through convolutional layers, enabling efficient feature extraction for tasks like semantic segmentation or motion prediction. Shallow CNN architectures have also been leveraged where they aim to make the learned kernel human interpretable. Despite this interpretability, they still require domain specific pretraining or fine-tuning~\cite{wilson2022convolutional}.}
    \item \textcolor{blue}{Recurrent Neural Networks (RNNs)~\cite{xiang2017darnn} incorporate temporal memory cells such as Long Short-Term Memory (LSTM) or Gated Recurrent Units (GRUs) to create statics scenes from sequential sensor observations.}
    
\end{itemize}
\textcolor{blue}{While neural methods demonstrate high performance, their reliance on implicit representations presents challenges for interpretability, and safety validation, particularly in mission-critical systems~\cite{wilson2022convolutional}. Furthermore, their architectures require retraining or fine-tuning for deployment across different domains or environments.}

% -------------------------------------------------
% Discuss the fundamentals of quasi-neural methods
% -------------------------------------------------
\subsubsection{Neuro-Symbolic Methods}
\textcolor{blue}{Neuro-Symbolic methods emerge at the intersection of traditional and neural methods, combining the computational efficiency of neural methods with the accuracy and interpretability of traditional methods.
Drawing from advances in cognitive science~\cite{plate2003holographic} and connectionist models~\cite{eliasmith2013build}, VSA-OGM introduces the \textbf{first neurosymbolic framework for occupancy grid mapping}. By leveraging biologically inspired hyperdimensional mathematics, VSA-OGM eliminates the need for domain-specific pretraining and does not require maintaining a dense map representation at runtime. This enables highly efficient, sparse updates and localized inferences, allowing it to scale to large environments while improving runtime latency compared to both traditional and neural methods.}

\begin{figure*}[h]
    \centering
    \includegraphics[width=0.8\textwidth]{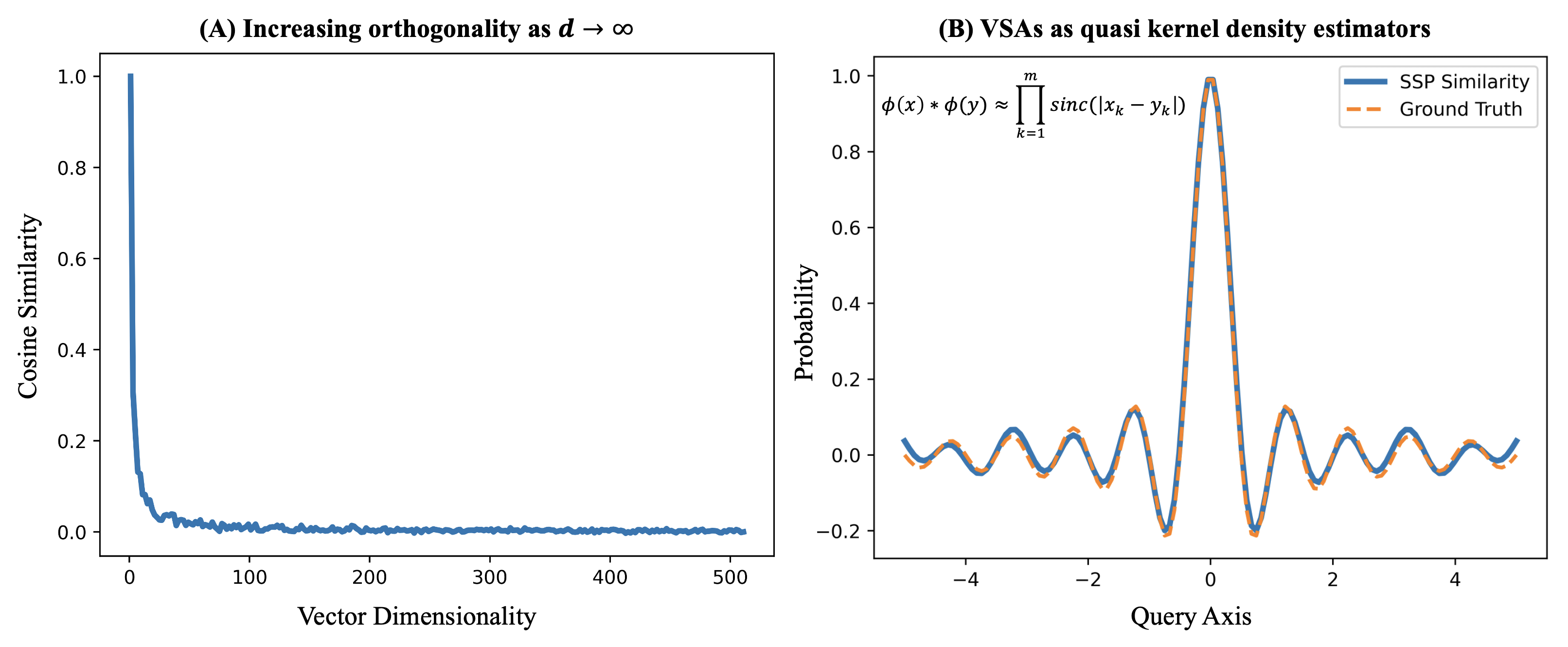}
    \caption{(A) The increasing orthogonality between pairs of randomly generated vectors as their dimensionality increases. (B) The dot product between two Spatial Semantic Pointers (SSPs) serving as a quasi kernel density estimator~\cite{voelker2020short}.}
    \label{fig:quasi-orth-kde}
\end{figure*}

\subsection{Vector Symbolic Architectures (VSAs)} \label{sec:vsa-background}

% ------------------------
% Discuss Origins of VSAs
% ------------------------
\textcolor{blue}{
VSAs present an algorithmic abstraction for the higher order computational properties of connectionist models~\cite{plate2003holographic}.
VSA systems leverage the quasi-orthogonality property that arises as the dimensionality of independent and identically distributed vectors approaches infinity \cite{plate1994distributed}.
}
\textcolor{blue}{
Figure~\ref{fig:quasi-orth-kde}A provides a visualization of this \textcolor{blue}{property} where quasi-orthogonality arises as vector dimensionality increases. The individual elements within hypervectors (HVs) depend on the specific type of VSA but generally consist of unit values~\cite{smolensky1990tensor, plate1995holographic}, complex unitary values~\cite{plate2003holographic}, or dense binary values~\cite{kanerva1994spatter}. This flexibility highlights the diversity of VSA literature where each has unique properties for specific applications~\cite{Kleyko_2022, kleyko2023survey}.
}
\textcolor{blue}{
VSAs utilize two primary operations: binding ($\otimes$) and bundling ($+$).
Binding, analogous to multiplication, combines HVs into a final invertible representation that is distinct from each input.
Bundling, analogous to addition, combines HVs into their superposition.
}

% -----------------------------------
% Discuss VSAs for Congitive Science
% -----------------------------------
\textcolor{blue}{
In this work, we adopt a specific class of VSAs based on unitary real-valued vectors known as Spatial Semantic Pointers (SSPs)~\cite{eliasmith2013build}, which are particularly well-suited for representing continuous spaces and performing deterministic neural computation. SSPs are grounded in Holographic Reduced Representations (HRRs)~\cite{plate1995holographic}, where binding is implemented via circular convolution and bundling through element-wise addition.}

\textcolor{blue}{SSPs go beyond HRRs with a specialized form of binding known as fractional binding~\cite{KomerBrent2020}, or fractional power encoding~\cite{plate1994distributed}, where real-valued spatial information is encoded by exponentiating real-valued vector elements in the complex domain. This mechanism allows SSPs to represent continuous-valued coordinates in a fixed-dimensional hypervector space.}
\textcolor{blue}{Furthermore, SSPs possess unique computational properties that support probabilistic encoding and inference~\cite{furlong2022fractional}. As shown in Figure~\ref{fig:quasi-orth-kde}B, the dot product of multiple SSPs produces a sinc function over the spatial domain, effectively performing quasi-kernel density estimation.
This process parallels methods such as Random Fourier Features~\cite{rahimi2007random} and Vector Function Architectures (VFAs)~\cite{frady2022computing}, where high-dimensional encodings are used to facilitate efficient kernel-based inference.}

\textcolor{blue}{Closely related to SSPs are Fourier Holographic Reduced Representations (FHRRs)~\cite{plate2003holographic}, another class of VSAs that perform operations in the complex domain to represent structured information in high-dimensional spaces. Prior work has used both SSPs and FHRRs to construct dynamic models of place cells for real-time spatial localization~\cite{frady2018framework, dumont2022model, dumont2025symbols}. While these efforts resemble the localization component of Simultaneous Localization and Mapping (SLAM)~\cite{durrant2006simultaneous}, they are restricted to sparse, point-based recognition and do not support dense spatial mapping or class-based region annotation, both of which are essential for OGM.}

\textcolor{blue}{A complementary line of research investigates VFAs and resonator networks for factorization~\cite{karunaratne2024rolenoisefactorizersdisentangling}, scene understanding~\cite{renner2024neuromorphic} and visual odometry~\cite{Renner2024-mi}. These systems typically rely on complete codebooks or pretrained visual templates for scene factorization, enabling recognition of previously seen spatial layouts. In contrast, OGM requires constructing dense, probabilistic maps from partial observations. For example, resonator-based odometry systems infer ego-motion from event-based sensors, whereas our goal is to probabilistically model high-fidelity spatial information from sparse sensor data.
Therefore, this work is, to the best of our knowledge, the first to apply VSAs toward the construction of large-scale, probabilistic OGMs in real-world environments.}

\section{Vector Symbolic Occupancy Grid Mapping} \label{sec:methods}

\textcolor{blue}{In Section~\ref{section:vsa-computation}, we discuss the mathematics for encoding statistical and spatial relationships within hyperdimensional space.}
In Section~\ref{section:framework-discussion}, we define our hyperdimensional OGM framework, \textcolor{blue}{Vector Symbolic Architectures for Occupancy Grid Mapping (VSA-OGM)}, and discuss how point cloud data is encoded, processed, and subsequently decoded \textcolor{blue}{to create probabilistic OGMs}.

{
\renewcommand{\arraystretch}{1.2}
\begin{table}[]
\begin{center}
\caption{The mathematical symbols used to describe VSA-OGM.} 
\label{tab:symbol-table}  
\begin{tabular}{c|c}
\hline
\textbf{Symbol}        & \textbf{Meaning}                  \\ \hline
$\otimes$              & circular convolution  \textcolor{blue}{(binding)}   \\
$+$                    & element-wise addition \textcolor{blue}{(bundling)}  \\
$*$                    & Dot product \\
$\odot$                & Hadamard product \\
$d$                    & vector dimensionality             \\
$n$                    & environmental dimensionality      \\
$\phi$                 & Spatial Semantic Pointer (SSP)    \\
$\phi_{axis}$          & \textcolor{blue}{axis SSP}                          \\
$\phi_{axis}(x)$       & $x\in\mathcal{R}$ encoded as a SSP \\
$\phi_{loc}$           & \textcolor{blue}{an n-d location encoded as an SSP} \\
$\mathcal{F}$          & Discrete Fourier Transform (DFT)  \\
$\mathcal{F}^{-1}$     & inverse DFT                       \\
$||\bullet||$          & vector magnitude                  \\
$l$                    & length scale                      \\
$\lambda$              & point cloud                       \\
$j$                    & number of points in each point cloud \\
$t$                    & time step                         \\
$\gamma$               & $n$-dimensional coordinate vector \\
$\psi$                 & \textcolor{blue}{class vector}                  \\
$\delta$               & number of \textcolor{blue}{tiles}  per dimension \\
$\omega$               & \textcolor{blue}{tile}  memory vector            \\
$\beta$                & singular \textcolor{blue}{tile}                 \\
$\mu$                  & center coordinate of $\beta$      \\
$r$                    & disk filter radius                \\
$M(x,y)$               & \textcolor{blue}{disk filter mask for $(x, y)$}     \\
$H$                    & Shannon entropy                   \\
$\rho$                 & decision threshold                \\
$\Omega$               & set of global memory vectors      \\
$N$                    & \textcolor{blue}{the number of agents} \\
$\Gamma$               & \textcolor{blue}{a single agent} \\
$Pr$                   & \textcolor{blue}{true probability value}            \\
$\mathcal{P}$          & \textcolor{blue}{quasi-probability value}           \\ 
$X$                    & \textcolor{blue}{set of discrete samples along $x$ axis} \\
$Y$                    & \textcolor{blue}{set of discrete samples along $y$ axis} \\ 
$d(x,y,x',y')$         & \textcolor{blue}{L2 distance between $(x,y)$ and $(x',y')$} \\
$OGM(x,y)$ & \textcolor{blue}{probabilistic signed difference at $(x, y)$} \\
$class(x, y)$ & \textcolor{blue}{inferred class of $(x, y)$ with $\rho$} \\
\hline
\end{tabular}
\end{center}
\end{table}
}

\subsection{Vector Symbolic Computation} \label{section:vsa-computation}

\noindent
We developed VSA-OGM based on Spatial Sementic Pointers (SSPs)~\cite{eliasmith2013build} for their unique probabilistic properties and ability to encode real-valued data. 
SSPs are defined as $d$-dimensional real-valued hypervectors normalized to unit length. 
All fundamental VSA operations are compatible with SSPs where grouping multiple vectors together, \textcolor{blue}{bundling, is achieved through element-wise addition} \textcolor{blue}{($+$)}:
\begin{equation} \label{eq:bundling}
    \phi(C) = \phi(A) + \phi(B).
\end{equation}
\textcolor{blue}{Binding uses circular convolution \textcolor{blue}{($\otimes$)} to associate vectors together, as shown in Equation~\ref{eq:binding}.}

\begin{equation}
\label{eq:binding}
\phi(A) \otimes \phi(B) = \mathcal{F}^{-1}\{\mathcal{F}\{ \phi(A) \} \odot \mathcal{F}\{ \phi(B) \} \},
\end{equation}
\textcolor{blue}{where $\mathcal{F}$ denotes the Fast Fourier Transform~\cite{Nussbaumer1981}. The resulting vector is orthogonal or nearly orthogonal to both inputs, enabling distinct compositional representations. This operation underlies the slot-filler representations commonly used in VSA systems~\cite{eliasmith2013build}.}

\textcolor{blue}{While many VSA systems~\cite{kleyko2023survey} represent discrete concepts using randomly sampled, mutually dissimilar vectors, they lack the ability to encode continuous spatial information. SSPs overcome this limitation by enabling fractional binding, which generalizes the standard binding operation to continuous domains. Specifically, SSPs allow real-valued spatial coordinates to be encoded by raising the Fourier transform of a base axis vector to a real-valued power~\cite{plate1995holographic, KomerBrent2020}:}

\begin{equation}
\label{eq:encoding-wout-ls}
\phi_{axis}(x) = \mathcal{F}^{-1}\{\mathcal{F}\{\phi_{axis}\}^x\}; \quad x \in \mathbb{R},
\end{equation}

\textcolor{blue}{This fractional binding mechanism is central to VSA-OGM, enabling continuous representations in high-dimensional space.}

\textcolor{blue}{Individual fractionally bound vectors} can be compared through a dot product operation. SSPs have a unique computational property where this dot product approaches a true sinc function as vector dimensionality $d$ approaches infinity~\cite{voelker2020short}. \textcolor{blue}{If} $x$ encoded as $\phi_{axis}(x)$ and the estimate of $x$ encoded as $\phi_{axis}(x)'$, the pseudo kernel density estimator  $k(x, x')$ is calculated as: $k(x, x') = \phi_{axis}(x) * \phi_{axis}(x)'$. A visualization of the decoded kernel is shown in Figure \ref{fig:quasi-orth-kde}B. The width of the induced kernel can be adjusted to specific values by modifying Equation \ref{eq:encoding-wout-ls} to
\begin{equation} \label{eq:fb-with-ls-1d}
        \phi\textcolor{blue}{_{axis}}(x) = \mathcal{F}^{-1}\{\mathcal{F}\{\phi\textcolor{blue}{_{axis}}\}^{(x/l)}\}; x \in \mathbb{R},
\end{equation}
where $l$ is the length scale parameter and there is a positive correlation between $l$ and the kernel width. This adjustment allows users to modulate the noise sensitivity of the estimator. Despite a lack of widespread use, sinc kernels have proven to be more accurate, in terms of mean integrated squared error, versus other kernels such as the Epanechnikov kernel~\cite{Tsybakov2009-fv}.

\begin{figure*}
    \centering
    \includegraphics[width=\textwidth]{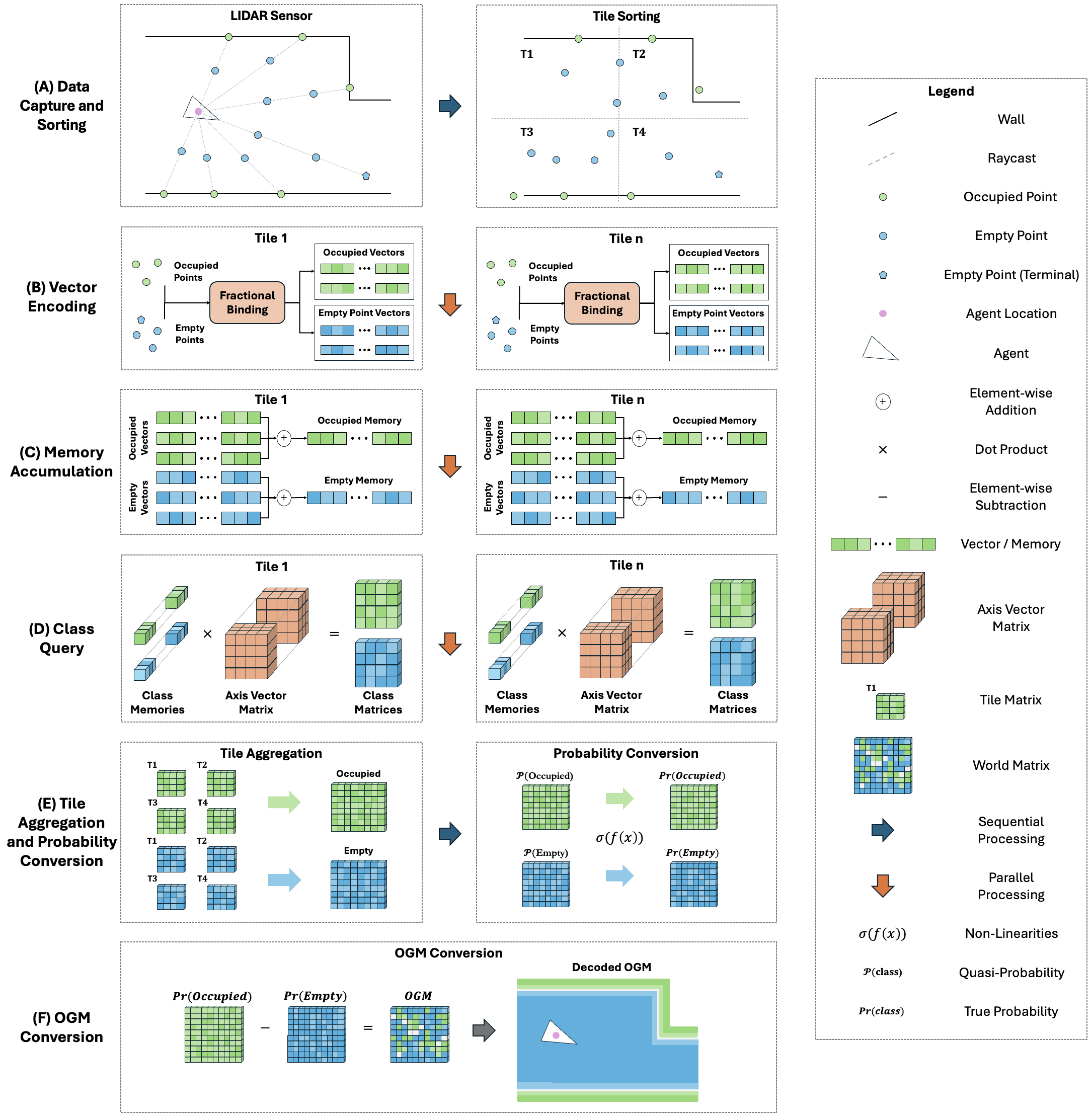}
    \caption{
        A high-level system overview of VSA-OGM. (A) Data capture and sorting with the training data being sampled from the LIDAR sensor before being sorted into individual tiles. (B) On a per-tile basis, the training data is fractionally bound into hyperdimensional vectors. (C) The vectors representing each data point are bundled into a single memory vector. (D) The aggregated class vectors are compared with the axis vector matrix, representing the tile's region encoded into hyperdimensional space, resulting in a class matrix that contains quasi-probability values representing the likelihood of each class at each discretized point. (E) The individual class matrices from each tile are aggregated together before being post processed with a series of non-linear post processing steps to remove noise and extract encoded information. (F) The individual class matrices are combined into a signal occupancy grid map with the probabilistic signed difference.
    }
    \label{fig:vsa-system-overview}
\end{figure*}

\subsection{Building Occupancy Grids} \label{section:framework-discussion}

\subsubsection{Data Format \& Preprocessing} OGM systems sample multiple point clouds $\{\lambda_0, \lambda_1, ..., \lambda_t\}$. For each timestep $t$, $\lambda_t$ consists of a coordinate vector $\gamma$ and a class vector $\psi$.
Since we are focusing on 2D OGM, all points clouds contain multiple 2D points representing x and y on the local reference frame. Moreover, the class vectors contain a single binary attribute signifying if a cell is occupied or empty.
To ensure fairness with the baseline methods, all point clouds are preprocessed to the global reference with the same simultaneous localization and mapping method described in~\cite{zhi2019continuous}.

\textcolor{blue}{A visualization of a labeled point cloud is provided in Figure~\ref{fig:vsa-system-overview}(A), where an agent samples the environment using a series of raycasts. Occupied points (green circles) are inferred from raycasts that terminate before reaching their maximum range, while points at the maximum range are labeled as empty (blue pentagons). Additional empty points are inferred by interpolating along the raycasts (blue points).}

\subsubsection{Initialization} Multiple properties of the VSA system must be specified before any point clouds can be processed.

\begin{itemize}
    \item \textit{Vector Dimensionality} $d$: The size of the individual vectors utilized within VSA-OGM.
    \item \textit{Length Scale} $l$: The width of the quasi-kernel induced by comparing multiple fractionally bound SSPs. We explore the impact of this parameter in Section~\ref{section:ablation-vector-length-scale}.
    \item \textit{Axis Vector}: The basis of fractionally binding Cartesian information in VSA systems. In the case of 2D OGM, we define axis vectors $\phi_{axis}^x$ and $\phi_{axis}^y$ representing the x and y axes. 
\end{itemize}

\subsubsection{Vector Discretization} \textcolor{blue}{Inspired by recent insights into chunking, where mammalian brains partition large scale spatial phenomena into distinct memory segments~\cite{chunking}, we abstracted this principle to discretize the environment into a matrix of tiles. This tiling enables VSA-OGM to learn multiple regions of the environment independently, a critical capability given that the bundling operation assumes all stored vectors are correlated. Without this separation, representing the entire environment with a single vector would weight all spatial regions equally, disregarding the distance dependent structure of the space.}

Specified by $\delta$, we split each dimension into $\delta$ evenly \textcolor{blue}{sized and spaced tiles}. To further discretize the environment, each \textcolor{blue}{tile} is also separated on a class-by-class basis. As such, for a given experiment with $\delta$ \textcolor{blue}{tiles} per dimension and $2$ classes, there are $\delta^n*2$ memory vectors \textcolor{blue}{$\omega$} with $d$ dimensionality:
\begin{equation}
    \omega_1, \omega_2, ..., \omega_{\delta^n * 2}.
\end{equation}
\textcolor{blue}{This process is demonstrated in Figure~\ref{fig:vsa-system-overview}(A), where the environment is discretized into four tiles to create spatially distinct training regions. Observed points within each tile are aggregated and processed in parallel, facilitating structured and scalable downstream learning.}

\subsubsection{Encoding Point Clouds}

\textcolor{blue}{Each point $\gamma_i$ in \textcolor{blue}{the point cloud} $\lambda_t$ is encoded into a set of location vectors \textcolor{blue}{${\phi_1, \phi_2, ..., \phi_j}$} by fractionally binding each coordinate of $\gamma_i$ along its respective dimension. This binding operation uses a variable length scale $l$ and is performed by expanding Equation~\ref{eq:binding}.}
\begin{equation} \label{eq:fb-with-ls-2d}
    \phi_{i} = \mathcal{F}^{-1}\{ \mathcal{F}\{ \phi_{axis}^x\}^{\gamma_i^x / l} \odot \mathcal{F}\{ \phi_{axis}^y\}^{\gamma_i^y / l} \}.
\end{equation}
This process is repeated for all $j$ points in $\lambda_t$ resulting in a set of $j$ location vectors representing each point in $\lambda_t$ encoded into hyperdimensional space. Figure~\ref{fig:vsa-system-overview}(B) visualizes this process, where occupied (green) and empty (blue) points are encoded by fractionally binding them with the axis basis vectors, producing the respective occupied and empty vectors for each observation.

\textcolor{blue}{Given a set of location vectors $\{\phi_1, \phi_2, \ldots, \phi_j\}$ and their corresponding classes $\{\psi_1, \psi_2, \ldots, \psi_j\}$ from $\lambda_t$, we follow Algorithm~\ref{alg:tile-mem-add} to add the $j$ points to the appropriate tile memory based on the closest tile and class. We begin by computing the center $\mu$ of each tile. Then, for each point, we identify the tile whose center is closest in Euclidean distance. Finally, we add the point’s vector $\phi_i$ to the memory associated with the nearest tile and label $\psi_i$.}
\textcolor{blue}{Figure~\ref{fig:vsa-system-overview}(C) visualizes this process, where occupied (green) and empty (blue) vectors are bundled into a single memory vector representing the cumulative representation of the observed point cloud. This process is also shown algorithmically in Algorithm~\ref{alg:tile-mem-add}.}

\begin{algorithm}
    \caption{Adding point vectors to \textcolor{blue}{tile} memories}\label{alg:tile-mem-add}
    \begin{algorithmic}
        \REQUIRE $\{\omega_{1}, \omega_{2}, ..., \omega_{\delta^n * 2}\}\gets$ \textcolor{blue}{tile} memory vectors
        \REQUIRE \textcolor{blue}{$\{\gamma_{1}, \gamma_{2}, ..., \gamma_{j}\} \gets$ points in Euclidean space}
        \REQUIRE $\{\phi_{1}, \phi_{2}, ..., \phi_{j}\} \gets$ location vectors
        \REQUIRE $\{\psi_{1}, \psi_{2}, ..., \psi_{j}\} \gets$ class vectors
        \REQUIRE $n \gets$ environment dimensionality
        \REQUIRE $\delta \gets$ \textcolor{blue}{tiles} per dimension
        \REQUIRE $\psi_i \in \{0, 1\}$
        \REQUIRE $n \geq 1$
        \REQUIRE $\delta \geq 1$
        \REQUIRE $j \geq 1$
        \STATE $i \gets 1$
        \STATE $\{\mu_1, ..., \mu_{\delta^n} \} \gets get\textcolor{blue}{Tile}Centers()$
        \WHILE{$i \leq j$}{
            \STATE $\beta \gets \arg \min\limits_{k} d_{L2}(\textcolor{blue}{\gamma}_i, \mu_k)$
            \STATE $\omega_{2\beta + \psi_i} \gets \omega_{2\beta + \psi_i} + \phi_i$
            \STATE $i \gets i + 1$
        }
        \ENDWHILE
    \end{algorithmic}
\end{algorithm}

\subsubsection{Decoding} \label{sec:decoding}

\textcolor{blue}{For \textcolor{blue}{OGM}, we focus on two key pieces of information: (1) which regions of the map are occupied, and (2) which regions are unoccupied or empty. Using the fully specified \textcolor{blue}{tile} memory vectors ${\omega_{1}, \omega_{2}, ..., \omega_{\delta^n*2}}$, we infer both occupancy and emptiness through deterministic vector operations combined with Shannon entropy~\cite{6773024}. This framework enables a probabilistic quantification of occupancy throughout the map. The decoding pipeline operates in four stages: (1) vector normalization, (2) similarity evaluation, (3) non-linear post-processing, and (4) entropy-based information extraction. We formally define and justify each of these steps in the following sections. The overall process is illustrated in Figure~\ref{fig:vsa-system-overview}(D–F). In Figure~\ref{fig:vsa-system-overview}(D), individual memory vectors are normalized and compared against the axis vector matrix to resolve class similarity. These tile-level results are then aggregated in Figure~\ref{fig:vsa-system-overview}(E), followed by non-linear post-processing to transform quasi-probabilities into valid probability values. Finally, Figure~\ref{fig:vsa-system-overview}(F) shows the combination of class probabilities to produce the resulting OGM.}

\textit{Vector Normalization:} Throughout the continuous operation of any SSP-based VSA system, successive bundling will slowly increase the magnitude of the memory vectors.
\textcolor{blue}{Therefore, a normalization operation transforming the memory vector back to unit length is required before querying any information}:
\begin{equation} \label{eq:vector-normalization}
    \omega_{i} = \frac{\omega_{i}}{\| \omega_{i} \|}.
\end{equation}

\textit{Similarity Evaluation:} \textcolor{blue}{After vector normalization, we} calculate the \textcolor{blue}{quasi-}probability \textcolor{blue}{$\mathcal{P}$ of a specific location $\gamma_i$ having a class of $\psi_i \in \{0, 1\}$ in a given tile }\textcolor{blue}{tile} $\beta$ as:
\textcolor{blue}{
\begin{equation} \label{eq:loc-query}
    \mathcal{P}(\psi_i|\gamma_i,\beta_i) = \omega_{2\beta + \psi_i} * \phi_{loc},
\end{equation}}
\textcolor{blue}{where $\phi_{loc}$ is a location vector} representing $\gamma_i$ in hyperdimensional space as described in Equation \ref{eq:fb-with-ls-2d}. The class of a region is calculated by expanding Equation \ref{eq:loc-query} across a matrix of $\phi_{loc}$. \textcolor{blue}{Depending on the specific application and the amount of available memory, the computational complexity of calculating $\phi_{loc}$ can be minimized by precalculating a matrix of location vectors and storing them in memory.} This process is shown visually in Figure~\ref{fig:vsa-system-overview}(D) where both class memory vectors are multiplied with the axis vector matrix to produce a matrix of quasi-probability values representing the likelihood of each class.

\textit{Non-linear Post-processing:}  
\textcolor{blue}{Even with vector normalization, the accumulation of noise from successive bundling renders direct decoding non-trivial. Many VSA frameworks address this challenge by employing clean-up memories or associative recall to project noisy vector representations onto more linearly separable factors~\cite{plate1995holographic, renner2024neuromorphic, Renner2024-mi, frady2018framework}. However, conventional codebook and resonator strategies are not applicable in the context of VSA-OGM. These approaches presuppose a fixed dictionary of factors and iteratively converge noisy vectors toward one of these predefined factors, after which an additional decoding step is required to obtain a real-valued location~\cite{Renner2024-mi, renner2024neuromorphic, frady2018framework}. By contrast, VSA-OGM requires well-defined similarity values across the entire spatial domain rather than the recovery of a single factor. Codebook- and resonator-based methods are therefore ill-suited to this task, whereas non-linear transformations applied directly to $\mathcal{P}$ yield more well-defined quasi-probability values.}

\textcolor{blue}{
We apply non-nonlinearities to $\mathcal{P}(\psi_i|\gamma_i,\beta_i)$ such that they are normalized by the maximum quasi-probability of each class and squared:
\begin{equation} \label{eq:non-linearities}
    \mathcal{P}(\psi_i|\gamma_i, \beta_i) = \left(\frac{\mathcal{P}(\psi_i|\gamma_i, \beta_i)}{\mbox{max}(\mathcal{P}(\psi_i))}\right)^2.
\end{equation}
We chose this combination of non-linearities based on the experimental analysis presented in Section~\ref{section:decoding method}. This combination of operations performs a normalization based on the maximum observed quasi-probability value and forces all values to be between 0 and 1 with the Born-rule from quantum probability theory~\cite{Born1926}.
}

\begin{algorithm}
    \caption{Calculating class-wise entropy}
    \begin{algorithmic}
        \REQUIRE $H(\psi) \gets$ matrix of global entropy values
        \REQUIRE $\mathcal{P}(\psi) \gets$ quasi-probability map of class $\psi$
        \REQUIRE $r(\psi) \gets$ disk filter radius for class $\psi$ 
        \REQUIRE $\psi \in \{0, 1\}$ 
        \REQUIRE $x \in \mathbb{N}$
        \REQUIRE $y \in \mathbb{N}$
        \STATE $x \gets 1$
        \STATE $y \gets 1$
        \WHILE{$x \leq \mathrm{numCells}(x)$}{
            \WHILE {$y \leq \mathrm{numCells}(y)$}{
                \STATE $m_1 = M(x, y, r(\psi=1))$ 
                \STATE $m_2 = M(x, y, r(\psi=0))$
                \STATE $H(\psi=1|x, y) = entropy(\mathcal{P}(\psi=1), x, y, m_1)$
                \STATE $H(\psi=0|x, y) = entropy(\mathcal{P}(\psi=0), x, y, m_2)$
                \STATE $y = y + 1$
            }
            \ENDWHILE
            \STATE $x = x + 1$
        }
        \ENDWHILE
    \end{algorithmic}
    \label{alg:entropy-filter}
\end{algorithm}

\textit{Entropy Information Extraction:} Shannon's theory of information has revolutionized modern communication systems and allows us to quantify information within noisy signals~\cite{6773024}. Recently, many works in material sciences~\cite{material-entropy} and radar imaging~\cite{9165582} use this well studied approach to extract feature rich information from noisy or otherwise incomprehensible data.
\begin{figure*}
    \centering
    \includegraphics[width=0.9\textwidth]{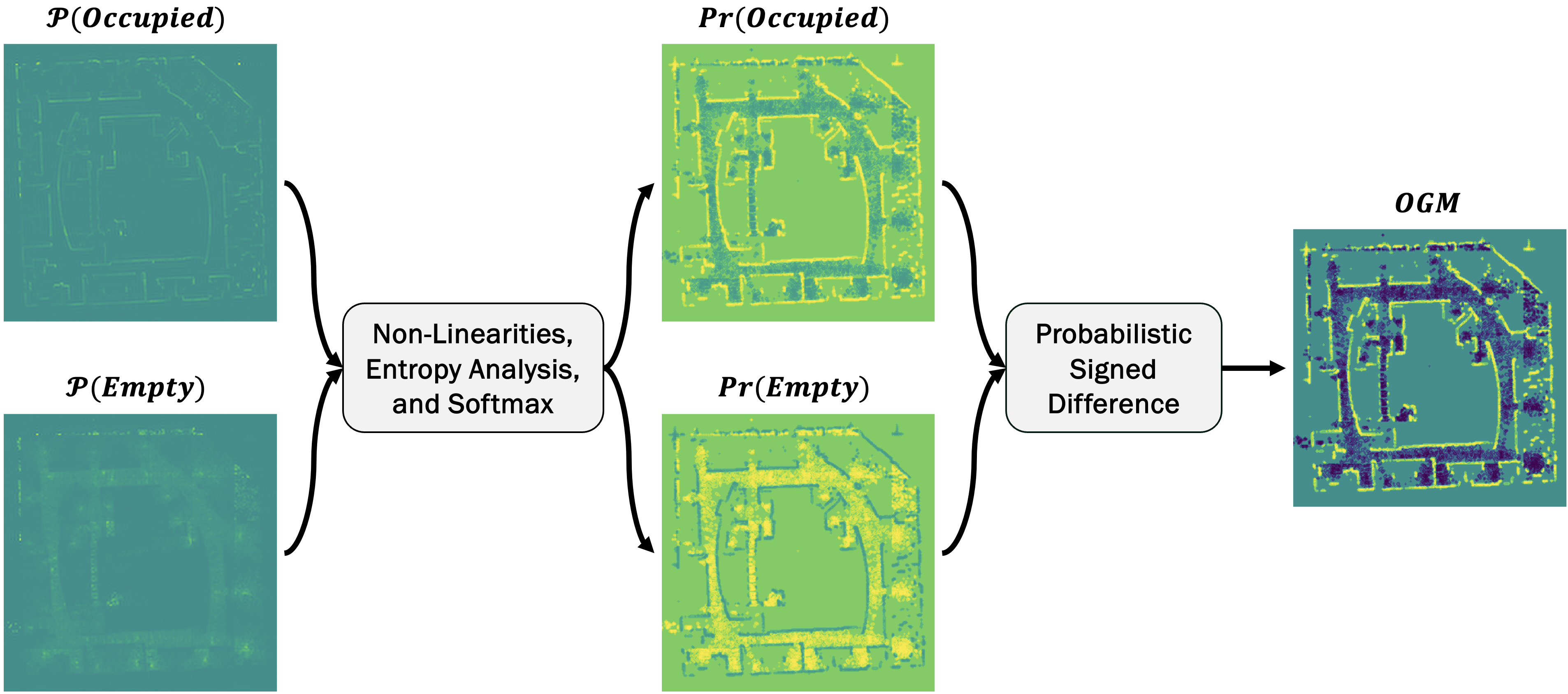}
    \caption{
        Visualization of VSA-OGM's post-processing pipeline, which transforms quasi-probability values $\mathcal{P}$ for each class into calibrated probabilities $Pr$ to generate the final occupancy grid map (OGM). The initial representation appears noisy or unclear because it directly reflects raw quasi-probabilities. Applying VSA-OGM’s non-linear decoding, entropy analysis, followed by softmax normalization and probabilistic signed difference, produces the clarified, class-separable dense OGMs.
    }
    \label{fig:entropy-calculation}
\end{figure*}
Taking inspiration from these works, we employ an entropy-based mechanism to extract labeled spatial information from the class matrices calculated by applying Equations \ref{eq:loc-query} \textcolor{blue}{and \ref{eq:non-linearities}} across the entire map \textcolor{blue}{resulting in $\mathcal{P}(\psi)$ that can be conditioned on discrete coordinates within the Cartesian plane.}

\textcolor{blue}{
We apply localized Shannon entropy defined as:
\begin{equation}
H(\psi|x, y) = -\sum_{x,y}^{X,Y} M(x, y) \mathcal{P}(\psi|x, y) \log \mathcal{P}(\psi|x, y),
\end{equation}
where $x \in X$ and $y \in Y$, with $X$ and $Y$ denoting the sets of discrete coordinates along the horizontal and vertical axes of the environment.
We define a binary mask function $M(x, y)$ such that it selects only the values within a radius $r$ of a given cell $(x, y)$.
The mask function is then defined as:
\begin{equation} \label{eq:entropy-mask}
M(x, y) =
\begin{cases}
1, & \text{if } d(x, y, x', y') \leq r \\
0, & \text{otherwise}
\end{cases},
\end{equation}
where $x'\in X$ and $y'\in Y$. Moreover, the Euclidean distance between the cell $(x, y)$ and its neighbor $(x', y')$ is defined as:
\begin{equation}
d(x, y, x', y') = \sqrt{(x' - x)^2 + (y' - y)^2}.
\end{equation}
}

Using \textcolor{blue}{masks} with radii $r(\psi=1)$ and $r(\psi=0)$ for both classes, this process is repeated across all \textcolor{blue}{discrete cells} resulting in class-wise \textcolor{blue}{entropy values} that increase class separability and reduce background noise.
\textcolor{blue}{We evaluate the performance benefits of our entropy-based decoding mechanism against more traditional activation functions in Section~\ref{section:ablation-entropy}.}
\textcolor{blue}{The specification of $r(\psi=1)$ and $r(\psi=0)$ and the implications thereof are explored in our ablation study in Section~\ref{section:ablation-entropy}. This process is shown programatically in Algorithm~\ref{alg:entropy-filter}.}

\begin{equation} \label{eq:softmax}
    Pr(\psi_i|x,y) = \frac{e^{H(\psi_i|x,y)}}{\sum_{j=1}^{2} e^{H(\psi_j|x,y)}}
\end{equation}
As shown in Equation~\ref{eq:softmax}, we apply the softmax function to convert the resulting class-wise entropy values to true probability values that sum to 1. This process is applied across all $x\in X$ and $y \in Y$. 
\begin{equation} \label{eq:psd}
    OGM(x, y) = Pr(\psi=1|x,y)-Pr(\psi=0|x,y)
\end{equation}
As shown in Equation~\ref{eq:psd}, we create our final map by applying the probabilistic signed difference to the individual class probabilities.
\textcolor{blue}{This entire process is highlighted visually in Figure~\ref{fig:entropy-calculation}, where the low fidelity quasi-probability values are post-processed into true probabilities with the combination of non-linearities, entropy analysis, and the softmax function. Lastly, the resulting OGM is created by using the probabilistic signed difference to aggregate the individual probability values into a single representation.}
A high-level visualization, within the context of the overall framework, is also shown in Figure~\ref{fig:vsa-system-overview}(E-F) where the quasi-probability class matrices are converted intro to true class probability matrices before being combined into a single OGM.
 
\textcolor{blue}{
With the final $OGM$ defined, we classify each point $(x, y)$ as:
\begin{equation} \label{eq:classification}
    \mathrm{class}(x, y) =
    \begin{cases}
        \mathrm{occupied}, & OGM(x,y) \geq \rho \\
        \mathrm{empty}, & else \\
    \end{cases},
\end{equation}
where $\rho$ is the decision threshold. The sensitivity of VSA-OGM to the proper specification of $\rho$ is highlighted with the area under the receiver operating characteristic curve~\cite{BRADLEY19971145} \textcolor{blue}{and the negative log-likelihood} in Section~\ref{sec:results}.
}

\subsubsection{Multi-Map Fusion}

\textcolor{blue}{VSA-OGM} supports multi-agent mapping through memory fusion. Given a set of $N$ independently operating agents $\{\Gamma_1, ..., \Gamma_N\}$, their \textcolor{blue}{tile} memory vectors, the hyperdimensional vectors storing each agent's observations, can be merged if and only if three conditions are satisfied. First, each agent $\Gamma_i$ must be operating on the same axis basis vectors for all dimensions of their operating environment. In this work focusing on two-dimensional \textcolor{blue}{OGM}, all agents must have the same axis basis vectors $\phi_{axis}^x$ and $\phi_{axis}^y$. Secondly, each agent should be learning the same number of classes with each having the same $\psi$ value. Lastly, each agent must use the same number of tiles per dimension $\delta$ to separate their environment into distinct \textcolor{blue}{tiles} along the global reference frame.

\begin{algorithm}
    \caption{Fusing multiple agent's memory vectors}
    \begin{algorithmic}
        \REQUIRE $\{\Gamma_1, ..., \Gamma_N\} \gets$ the set of all agents
        \REQUIRE $N \geq 2$
        \STATE $\Omega \gets \emptyset$
        \STATE $i \gets 1$
        \WHILE{$i \leq N$}{
            \STATE $j \gets 1$
            \WHILE {$j \leq getNumMemories(\Gamma_i)$}{
                \STATE $\omega_j = getMemory(\Gamma_i, j)$
                \IF {$\omega_j \notin \Omega$}
                    \STATE $\Omega = \Omega \cup \omega_j$ \COMMENT{Add $\omega_j$ to the set $\Omega$}
                \ELSE
                    \STATE $\Omega\{\omega_j\} = \Omega\{\omega_j\} + \omega_j$
                \ENDIF

                \STATE $j = j + 1$
            }
            \ENDWHILE
            \STATE $i = i + 1$
        }
        \ENDWHILE
    \end{algorithmic}
    \label{alg:fusion}
\end{algorithm}

\textcolor{blue}{Assuming all three conditions are met, their unique tile memories can be merged into a cohesive set of global tile memories, representing the collective knowledge amongst the group. The process begins by initializing an empty set $\Omega$ that will contain the unique tile memory vectors within all agents. Now iterating over each agent, the following steps are followed. First, check if all \textcolor{blue}{unique  tiles} within the agent are represented within $\Omega$. Second, if all tiles exist within $\Omega$, proceed to the third step. Otherwise, empty tile memory vectors are initialized within $\Omega$ corresponding to the unique tile within the current agent. Third, use Equation \ref{eq:bundling}, to bundle each tile memory vector within the current agent with the global tile memory vector within $\Omega$. Lastly, repeat this process until all agents and all tile memories have been accounted for in $\Omega$. This approach is also shown with psuedocode in Algorithm~\ref{alg:fusion}.}
{
\renewcommand{\arraystretch}{1.2}
\begin{table*}[]
\centering
\caption{\textcolor{blue}{A summary of all single-agent datasets used in this paper: Ablation (Dense), Ablation (Sparse), ToySim~\cite{senanayake2017bayesian}, Intel Map~\cite{Radish}, and EviLOG~\cite{evilog}.Dataset density is defined as the total number of labeled points divided by the environment area (points/m²). (–) Points are provided without labels.}}
\label{table:data-summary}
\begin{tabular}{c|c|c|c|c|c|c|c}
\hline
\textbf{Dataset}                              & \textbf{Type}     & \textbf{Map Size} & \textbf{Points / Scan} & \textbf{\# Scans} & \textbf{\% Occupied} & \textbf{\% Empty} & \textbf{Density} \\ \hline
\textbf{Ablation (Dense)}                     & Simulated & $36\mbox{m}^2$    & 100     & 100   & $50\%$ & $50\%$ & $277.8$ \\
\textbf{Ablation (Sparse)}                    & Simulated & $36\mbox{m}^2$    & 100     & 10    & $50\%$ & $50\%$ & $27.8$  \\
\textbf{ToySim}~\cite{senanayake2017bayesian} & Simulated & $10000\mbox{m}^2$ & $50$    & $90$  & $60\%$ & $40\%$ & $0.45$  \\
\textbf{Intel Map}~\cite{Radish}              & Real      & $1440\mbox{m}^2$  & $421$   & $910$ & $58\%$ & $42\%$ & $266$   \\
\textbf{EviLOG}~\cite{evilog}                 & Simulated & $4614\mbox{m}^2$  & $48000$ & 11100 & -      & -      & -       \\ \hline
\end{tabular}
\end{table*}
}

{
\renewcommand{\arraystretch}{1.2}
\begin{table*}[]
\centering
\caption{\textcolor{blue}{A summary of all multi-agent datasets used in this paper: ToySim~\cite{senanayake2017bayesian} and Intel Map~\cite{Radish}.}}
\label{table:ma-data-summary}
\begin{tabular}{c|c|c|c|c|c|c}
\hline
\textbf{Dataset}                                               & \textbf{Agent Id} & \textbf{Map Size}                  & \textbf{Points / Scan} & \textbf{\# Scans}    & \textbf{\% Occupied} & \textbf{\% Empty} \\ \hline
\multirow{2}{*}{\textbf{ToySim}~\cite{senanayake2017bayesian}} & 1                 & \multirow{2}{*}{$10000\mbox{m}^2$} & \multirow{2}{*}{$50$}  & \multirow{2}{*}{$367$} & $48.9\%$             & $51.1\%$          \\
                                                               & 2                 &                                    &                        &                        & $53.1\%$             & $46.9\%$          \\ \hline
\multirow{4}{*}{\textbf{Intel Map}~\cite{Radish}}              & 1                 & \multirow{4}{*}{$1440\mbox{m}^2$}  & \multirow{4}{*}{$421$}                       & 222                    & $38.8\%$            & $61.2\%$         \\
                                                               & 2                 &                                    &                        & 161                    & $43.4\%$            & $56.6\%$         \\
                                                               & 3                 &                                    &                        & 283                    & $42.3\%$            & $57.7\%$         \\
                                                               & 4                 &                                    &                        & 244                    & $42.1\%$            & $57.9\%$         \\ \hline
\end{tabular}
\end{table*}
}

\section{Results} \label{sec:results-intro}

To evaluate the performance of VSA-OGM, we use \textcolor{blue}{five} datasets and \textcolor{blue}{nine} baseline methods. The first dataset was synthetically created with the simulator from Bayesian Hilbert Maps~\cite{senanayake2017bayesian} (\textit{ToySim}). The second is the Intel Campus dataset that was captured with a physical robot at Intel's Seattle campus~\cite{Radish} (\textit{Intel Map}). The third is the simulated autonomous driving dataset from Evidential LiDAR Occupancy Grid Mapping (EviLOG)~\cite{evilog} (\textit{EviLOG}). The \textcolor{blue}{fourth and fifth are} simplified datasets we created for our ablation study \textcolor{blue}{with dense (\textit{AblationDense}) and sparse (\textit{Ablation-Sparse}) variants. A summary of these datasets is shown in Table~\ref{table:data-summary} with all multi-agent variants shown in Table~\ref{table:ma-data-summary}.}
Mapping accuracy is measured as the area under the receiver operating characteristic curve (AUC)~\cite{BRADLEY19971145} \textcolor{blue}{and the negative log-likelihood (NLL). All experiments are conducted using a desktop workstation with an Intel(R) Xeon(R) W-2295 CPU, 128GB RAM, and an RTX A5000 GPU}.
We split the overall results into two sections.
In Section \ref{sec:results}, we compare our approach with \textcolor{blue}{8} traditional methods and \textcolor{blue}{one neural method}. \textcolor{blue}{Since most existing neural OGM methods are designed for 3D environments~\cite{wilson2022convolutional, schreiber2022multi, zhang2024rapid}, we restricted our comparison to a 2D neural baseline to ensure a fair and direct evaluation against our 2D approach.} \textcolor{blue}{A summary of all parameters used in each experiment is shown in Table~\ref{tab:hyperparameters}.}
In Section \ref{sec:ablation}, we perform an ablation study of VSA-OGM across \textcolor{blue}{our entropy based decoding method, the environmental tiling parameter}, the disk filter radii parameter, and the vector length scale parameter.

{
\renewcommand{\arraystretch}{1.2}
\begin{table*}[]
\centering
\caption{\textcolor{blue}{A summary of the VSA-OGM parameters used in all experiments.}}
\label{tab:hyperparameters}  
\begin{tabular}{c|ccccc|c}
\hline
\textbf{Experiment}                                             & $d$   & $l$ & $\delta$ & $r(\psi=0)$ & $r(\psi=1)$ & \textbf{Multi Agent?} \\ \hline
\multirow{2}{*}{\textbf{ToySim}~\cite{senanayake2017bayesian}} & \multirow{2}{*}{32000} & \multirow{2}{*}{1.0} & \multirow{2}{*}{2}        & \multirow{2}{*}{3}           & \multirow{2}{*}{3}           & \ding{55}             \\
                                                                &       &     &          &             &             & \checkmark            \\ \hline
\multirow{2}{*}{\textbf{Intel Map}~\cite{Radish}}               & \multirow{2}{*}{32000} & \multirow{2}{*}{0.2} & \multirow{2}{*}{8}        & \multirow{2}{*}{1}           & \multirow{2}{*}{1}           & \ding{55}             \\
                                                                &       &     &          &             &             & \checkmark            \\ \hline
\textbf{EviLOG}~\cite{evilog}                                   & 32000  & 0.2 & 1         & 3            & 3            & \ding{55}             \\ \hline
% \textbf{nuScenes}~\cite{caesar2020nuscenes}                     &       &     &          &             &             & \ding{55}             \\ \hline
\end{tabular}
\end{table*}
}

% -------------------------------
% Comparing VSA versus Baselines
% -------------------------------
\subsection{VSA versus Baselines} \label{sec:results} 

Using the aforementioned datasets, we split our experiments into \textit{Single Agent} and \textit{Multi Agent} tasks, where the former uses a single agent to map the entire environment and the latter uses multiple agents whose individual perspectives are combined to create a single ``fused" \textcolor{blue}{occupancy grid map (OGM)}.

% -----------------------------------------
% Section: Single Agent Simulated Mapping
% -----------------------------------------
\subsubsection{ToySim - Single Agent} \label{sec:sim-data-single}

We created a synthetic LiDAR dataset with a re-implementation of the simulator introduced in BHM~\cite{Senanayake_Ramos_2018}. This dataset uses a single agent equipped with a single LiDAR sensor having a $360^{\circ}$ field of view, a view distance of $20$ meters, and $50$ individual beams. The total area of the environment is $10000\mbox{m}^2$. The dataset contains $4500$ individual points with a label distribution of $60\%$ occupied and $40\%$ empty. \textcolor{blue}{The dataset statistics are also summarized in Table~\ref{table:data-summary}.}

We train VSA-OGM and BHM~\cite{senanayake2017bayesian} on $80\%$ of the dataset with $20\%$ reserved for validation where the \textcolor{blue}{final AUC scores and negative log likelihood values are determined.} This dataset is particularly difficult due to the sparse features. Many traditional applications of probabilistic \textcolor{blue}{OGM} avoid this issue by augmenting the training data. This is accomplished by interpolating a linear line with random points between the agent's location and the LiDAR points~\cite{senanayake2017bayesian}. We chose to avoid this preprocessing step to highlight how \textcolor{blue}{traditional} methods and VSA-OGM respond to data sparsity. 

{
\renewcommand{\arraystretch}{1.2}
\begin{table*}[]
\begin{center}
\caption{A comparison between VSA-OGM versus Bayesian Hilbert Maps (BHM)\cite{Senanayake_Ramos_2018}, \textcolor{blue}{with and without a fully defined covariance matrix,} on the simulated dataset created from our reimplementation of the simulator introduced in \cite{senanayake2017bayesian}. \textcolor{blue}{($-$) open-source implementation is unavailable.}}
\label{tab:simulated-single-agent-results}  
\begin{tabular}{c|cc|cc|c}
\hline
\textbf{Method}                                   & \textbf{AUC} & \textbf{NLL} & \textbf{CPU Latency} & \textbf{GPU Latency} & \textbf{Model Size} \\ \hline
\textbf{BHM - Full}~\cite{senanayake2017bayesian} & 0.96         & 0.44         & 6.5s              & $-$           & $6400$ MB           \\
\textbf{BHM - Diag}~\cite{senanayake2017bayesian} & 0.97         & 0.38         & 0.45s             & 0.024s            & $0.04$ MB         \\
\textbf{VSA-OGM (ours)}                           & 0.93         & 0.33         & 0.04s             & 0.005s            & $1.02$ MB          \\ \hline
\end{tabular}
\end{center}
\end{table*}
}

\begin{figure*}
    \centering
    \includegraphics[width=0.85\textwidth]{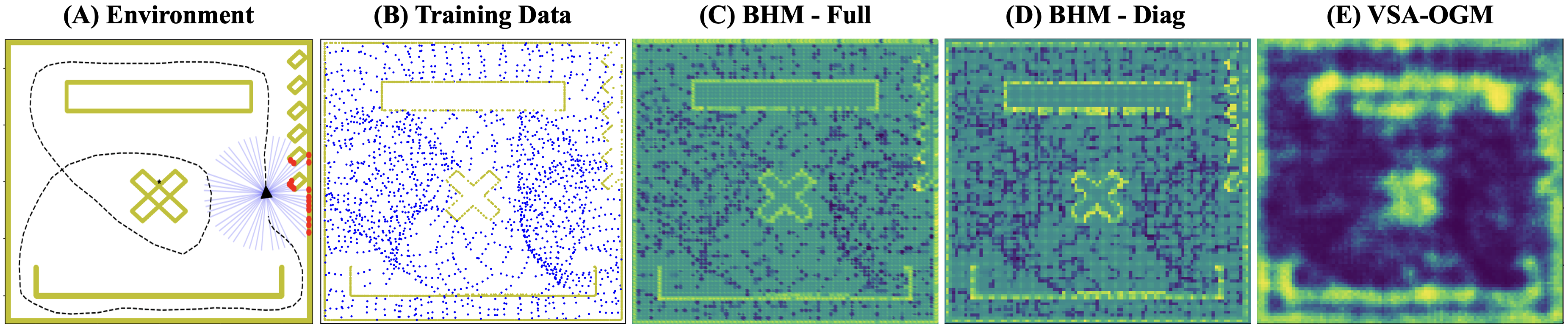}
    \caption{Simulated single agent mapping experiment with the dataset generated from the simulator introduced in Bayesian Hilbert Maps (BHM)~\cite{senanayake2017bayesian}. (A) The simulated testing environment with the agent's trajectory. (B) The sparse point cloud data with ground truth labels for training. \textcolor{blue}{(C) Trained OGM from BHM with a fully-defined covariance matrix. (D) Trained OGM from BHM with a diagonal covariance matrix. (E) Trained OGM from VSA-OGM.}}
    \label{fig:simulated-single-agent-results}
\end{figure*}

With a vector dimensionality of $32000$, a length scale of $1.0$, occupied and empty disk filter radii of $3$ \textcolor{blue}{cells}, and $2$ tiles per dimension, VSA-OGM achieves an AUC score of $0.93$ versus BHM with scores of $0.96$ \textcolor{blue}{and $0.97$ for the versions with a full and diagonal covariance matrix, respectively. BHM - Full maintains covariance terms between all environmental cells, whereas BHM - Diag discards  covariance (retaining only per-cell variances). Comparing the resulting NLL scores, VSA-OGM had the lowest value of 0.33 compared to BHM - Diag with a value of 0.38 and BHM - Full with a value of 0.44. \textcolor{blue}{Although the simultaneous improvement in NLL and reduction in AUC may appear counterintuitive, this indicates that VSA-OGM is less effective at ranking data points but produces probability estimates that are better calibrated to the true values.} This characteristic of VSA-OGM is a result of the utilization of a softmax function to convert the quasi-probability values into true values. Refer to Equation~\ref{eq:softmax} and Figure~\ref{fig:entropy-calculation}(B) for more information.}

Despite adjusting the kernel-width parameter for either BHM implementation, we were unable to build an \textcolor{blue}{OGM} that successfully interpolates between sparse points to broadly classify empty regions with no training or testing data. \textcolor{blue}{Therefore, all results presented in this section use the same hyperparameters as specified in Section~\ref{sec:intel-global}.} This is shown in Figure~\ref{fig:simulated-single-agent-results} where the majority of the map is \textcolor{blue}{labeled} as unknown, highlighted by the teal color. On the other hand, VSA-OGM was able to successfully interpolate between the empty points and label the majority of the area as empty.
Our increased interpolation performance also comes with significantly reduced inference latency at approximately \textcolor{blue}{162}x (CPU) and \textcolor{blue}{1300x} (GPU) reductions versus BHM - Full (CPU) while maintaining comparable accuracy on the training and validation sets.
Our approach reduces the total memory size of the learned representation by \textcolor{blue}{6400x}. \textcolor{blue}{The GPU latency numbers for BHM - Full are omitted because we were unable to find a working GPU-compatible implementation.}  
\textcolor{blue}{Compared to BHM - Diag, VSA-OGM provides latency reductions of approximately 10x (CPU) and 5x (GPU) at the expense of increased memory complexity. However, we still see the same issue where neither version of BHM is able to interpolate between the sparsely observed regions of the environment. This is shown visually in Figure~\ref{fig:simulated-single-agent-results}(C-D), where both BHM variants produces sharp, high-frequency peaks that closely follow the training data but provide little interpolation in sparse regions. In contrast, VSA-OGM generates coarser, lower-frequency features that, while less detailed, interpolate smoothly across unlabeled regions with reduced latency.} These results are summarized in Table~\ref{tab:simulated-single-agent-results}. The labeled points and trained \textcolor{blue}{OGMs} are visualized in Figure~\ref{fig:simulated-single-agent-results}.

\subsubsection{ToySim - Multi-Agent} \label{sec:sim-data-fusion}

{
\renewcommand{\arraystretch}{1.2}
\begin{table}[]
\begin{center}
\caption{The AUC score~\cite{BRADLEY19971145}, Negative Log-Likelihood (NLL), and the per point cloud processing of agents 1 and 2 on our simulated dataset before being merged into a single, fused representation.}
\begin{tabular}{l|cc|c}
\hline
\textbf{Agent}       & \textbf{AUC} & \textbf{NLL} & \textbf{Latency (s)} \\ \hline
\textbf{Agent 1}     & $0.93$       & $0.32$       & $0.005$              \\
\textbf{Agent 2}     & $0.96$       & $0.46$       & $0.005$              \\
\textbf{Map Fusion}  & $0.95$       & $0.32$       & $0.006$              \\ \hline
\end{tabular}
\label{tab:sim-fusion-results}  
\end{center}
\end{table}
}

We created a second dataset from our re-implemented simulation environment from \cite{Senanayake_Ramos_2018}. Rather than being focused on a single agent, this experiment simulates the accumulation of data from two independent agents operating within the same environment. As such, each will have a unique perspective of the environment that should help improve the collective world model.
\begin{figure*}
    \centering
    \includegraphics[width=0.65\textwidth]{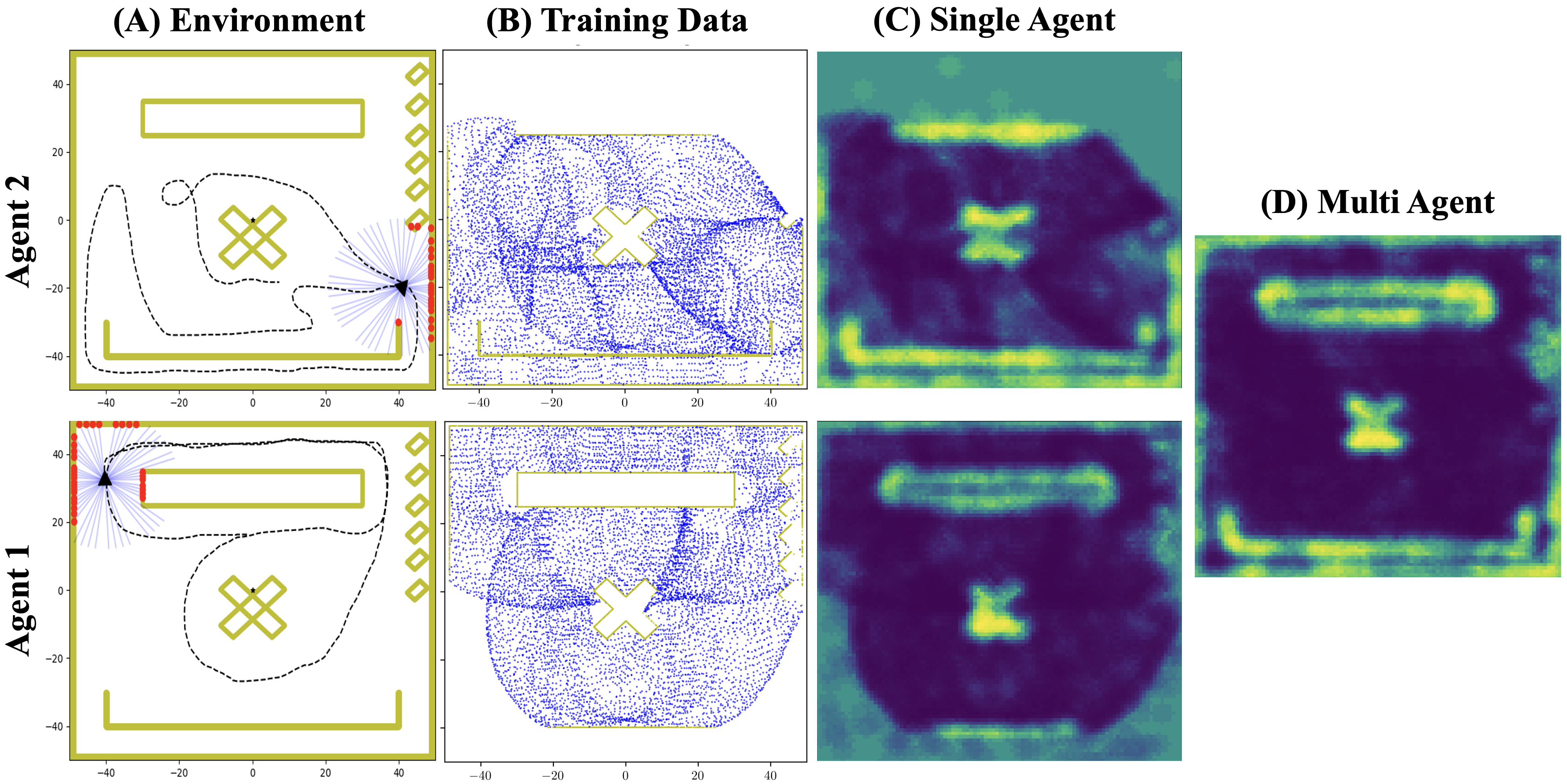}
    \caption{Simulated multi-agent mapping experiment. \textcolor{blue}{(A) The simulated testing environments with both agent's unique trajectory (B) The sparse point cloud data with ground truth labels for training. (C) The global combined OGM from VSA-OGM across each individual agent. (D) The combined OGM from VSA-OGM with the combined knowledge from Agent 1 and Agent 2.}}
    \label{fig:simulated-fusion-maps-individual}
\end{figure*}
The same configuration parameters are used from Section \ref{sec:sim-data-single}. Therefore, both agents are equipped with identically configured hyperdimensional mapping systems. Both agent's are trained simultaneously on $80\%$ of their training data along with fusion occurring at every time step. The AUC score is calculated across both agent's individually with Agent 1 achieving \textcolor{blue}{an AUC score of $0.93$ and a NLL of $0.32$. Agent 2 achieves an AUC score of $0.96$ and a NLL of 0.46.} Furthermore, the final fused map achieves \textcolor{blue}{an AUC score of $0.95$ and a NLL of $0.32$}, suggesting that our fusion methodology results in negligible information losses and produces a global representation with high AUC values and low NLL scores. The individual training results amongst both agents are shown in Figure \ref{fig:simulated-fusion-maps-individual}. \textcolor{blue}{The fusion operation is approximately equivalent to processing one point cloud and generating the map at 5ms. These performance and latency numbers for each agent and the fused representation are shown in Table~\ref{tab:sim-fusion-results}.}

{
\renewcommand{\arraystretch}{1.2}
\begin{table*}[]
\centering
\caption{\textcolor{blue}{Comparison between VSA-OGM, OGM~\cite{elfes1989using}, I-GPOM~\cite{Ghaffari_Jadidi2018-vg}, I-GPOM2~\cite{Ghaffari_Jadidi2018-vg}, SBHM~\cite{senanayake2017bayesian}, Fast-BHM~\cite{zhi2019continuous}, and SBKM~\cite{duong2022autonomoussbkm} on the Intel campus dataset~\cite{Radish} (space delta) the resolution of the decoded OGM in meters per cell. (*) additional experiments we performed with the baseline methods. (\ding{55}) unable to evaluate due to excessive GPU memory usage. ($\circ$) values are not listed in original manuscript. ($\dagger$) Original manuscript didn't specify if latency values include point cloud processing and map generation.}}
\label{tab:intel-single}
\begingroup
\renewcommand{\arraystretch}{1.5}
\begin{tabular}{c|cccc|cc|c|cc}
\hline
\textbf{Parameters}                                           & $\gamma$ & $\Sigma$        & \# features & space delta & AUC    & NLL     & Model Size & CPU Time & GPU Time    \\ \hline

\textbf{OGM~\cite{elfes1989using}}                            & -        & -               & -           & $\circ$     & 0.93   & $\circ$ & $\circ$    & 1.12s    & -           \\ 
\textbf{I-GPOM~\cite{Ghaffari_Jadidi2018-vg}}                 & 3/2      & full            & -           & $\circ$     & 0.94   & $\circ$ & $\circ$    & 15.35s   & -           \\ 
\textbf{I-GPOM2~\cite{Ghaffari_Jadidi2018-vg}}                & 3/2      & full            & -           & $\circ$     & 0.97   & $\circ$ & $\circ$    & 17.17s   & -           \\ \hline
\multirow{4}{*}{\textbf{BHM}~\cite{senanayake2017bayesian}}   & 6.71     & full            & \multirow{4}{*}{5600}     & 0.1    & 0.96    & $\circ$    & 6400MB   & 22.35s       & - \\
                                                              & 6.71     & diag            &             & 0.1         & 0.96*  & 0.25*   & 0.04 MB    & 8.11s*   & \ding{55}*  \\
                                                              & 6.71     & full            &             & 0.2         & 0.98   & 0.24    & 6400 MB    & 11.8s*    & -           \\
                                                              & 6.71     & diag            &             & 0.2         & 0.98   & 0.24    & 0.04 MB    & 2.11s*    & 0.11s*      \\ \hline
\textbf{Fast-BHM~\cite{zhi2019continuous}}                    & $\circ$  & diag            & 5600        & 0.1         & 0.95   & $\circ$ & 0.04 MB    & 0.46s$^\dagger$    & -           \\ \hline
\multirow{2}{*}{\textbf{SBKM~\cite{duong2022autonomoussbkm}}} & 6.71     & full            & \multirow{2}{*}{3492}     & 0.2    & 0.96    & 0.36       & $\circ$  & 0.76s$^\dagger$    & -           \\
                                                              & 6.71     & $\lambda_{max}$ &             & 0.2         & 0.95   & 0.52    & $\circ$    & 0.43s$^\dagger$    & -           \\ \hline
\multirow{2}{*}{\textbf{VSA-OGM}}                             & -        & -               & \multirow{2}{*}{32000}    & 0.1    & 0.95    & 0.25       & \multirow{2}{*}{16.3MB} & 0.51s    & 0.04s       \\
                                                              & -        & -               &             & 0.2         & 0.95   & 0.25    &            & 0.26s    & 0.02s       \\ \hline
\end{tabular}
\endgroup
\end{table*}
}

\subsubsection{Intel Map - Single Agent} \label{sec:intel-global}

\begin{figure}
    \centering
    \includegraphics[width=0.75\linewidth]{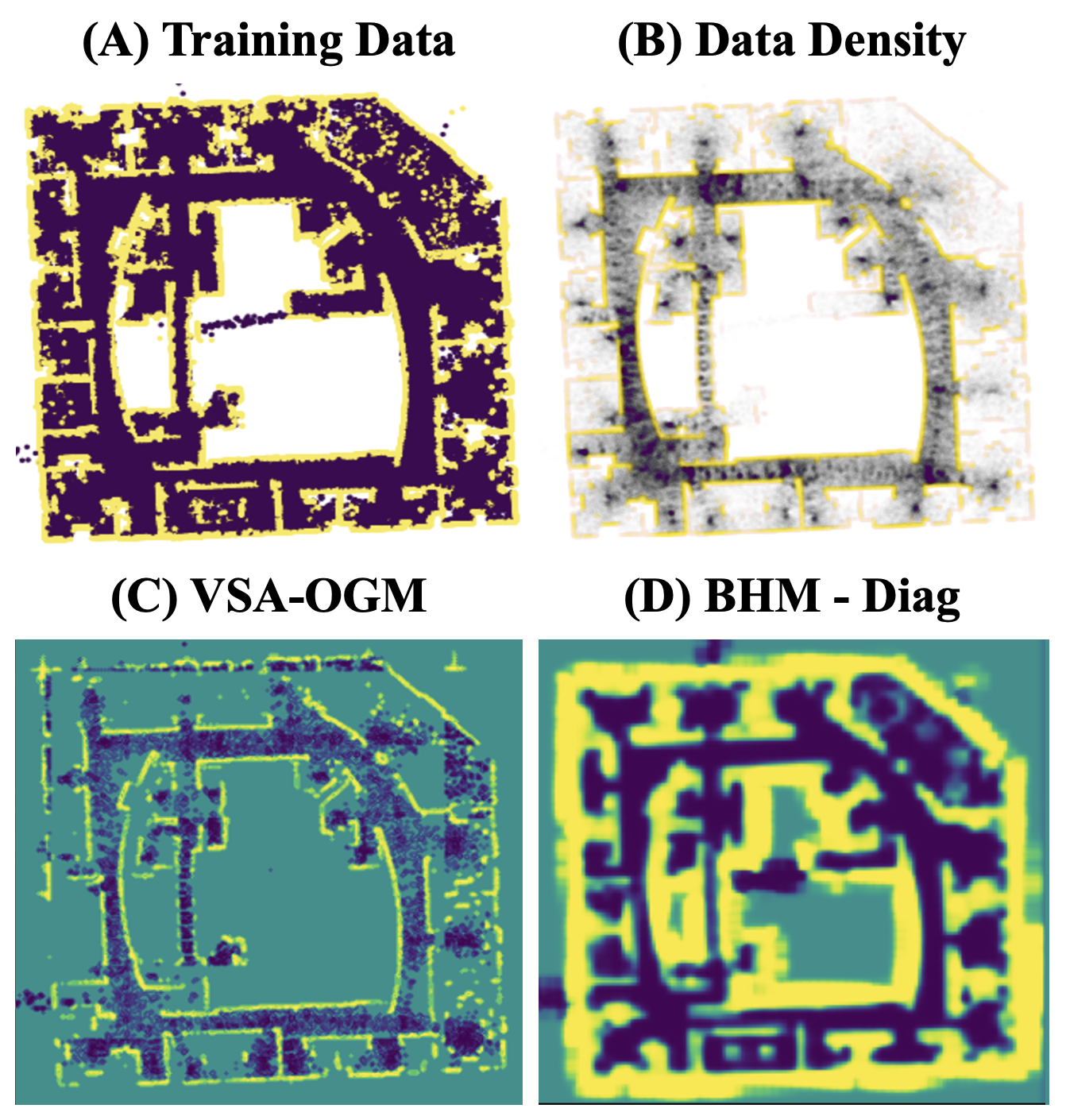}
    \caption{Single agent mapping experiment with the Intel Map dataset~\cite{Radish}: (A) The ground-truth labeled training data. \textcolor{blue}{(B) The density of the training data. (C) The global combined OGM from VSA-OGM. (D) The final OGM from BHM with a diagonal covariance matrix.}}
    \label{fig:intel-single}
\end{figure}

The Intel \textcolor{blue}{Map} dataset \cite{Radish} is a commonly used, real-word dataset for benchmarking \textcolor{blue}{OGM} algorithms. Using a vector dimensionality of \textcolor{blue}{$32000$}, a length scale of $0.2$ meters, disk filter radii of $1$ cell, and \textcolor{blue}{$8$} quadrants per dimension, \textcolor{blue}{VSA-OGM (CPU)} achieves an AUC score of $0.95$ with an average per point cloud processing time of $510$ms versus BHM, with a full covariance matrix (BHM - Full),~\cite{senanayake2017bayesian} with a score of $0.96$ and a processing time processing time of $22.35$s.
% \textcolor{blue}{More information on the derivation of VSA-OGM's parameters can be found in our ablation study in Section~\ref{sec:ablation}.}
Furthermore, \textcolor{blue}{BHM with a diagonal covariance matrix (BHM - Diag) achieves a final AUC of 0.96 with a processing time of 8.11s.} These results are summarized in Table \ref{tab:intel-single} with the ground truth representations and trained occupancy grid maps visualized in Figure \ref{fig:intel-single}. 

We were unable to compare against BHM - Full on the GPU because the open-source implementation does not support GPU execution. BHM - Diag does support GPU execution, and VSA-OGM achieves an order-of-magnitude latency reduction at a spatial resolution of 0.2m. However, we could not scale BHM - Diag to a resolution of 0.1m because its dynamic memory requirements exceeded the available 24GB of GPU memory. In terms of memory, VSA-OGM provides an approximate 392x reduction compared to the BHM - Full. No memory reduction is achieved with respect to BHM - Diag because this method assumes indepenedence between cells, which drastically lowers the memory footprint.

Relative to Fast-BHM~\cite{zhi2019continuous}, VSA-OGM reduces per point cloud latency by an order of magnitude while maintaining covariance between cells. Unfortunately, a GPU-compatible implementation of Fast-BHM was unavailable, and the runtimes reported in the original paper were unclear as to whether they included both map generation and training. If map generation was included (as in our BHM experiments) its latency would likely be substantially higher than the reported numbers. Since Fast-BHM assumes independence between voxels, no memory reduction is seen compared to VSA-OGM.

As shown in Table~\ref{tab:intel-single}, VSA-OGM offers one to two orders of magnitude latency reductions compared to earlier traditional methods~\cite{elfes1989using, Ghaffari_Jadidi2018-vg}. Compared to a more recent traditional approach~\cite{duong2022autonomoussbkm}, VSA-OGM reduces CPU latency by approximately 2x.

VSA-OGM establishes a new paradigm in occupancy grid mapping: a neurosymbolic approach that is faster, more memory-efficient, and extensible to future scaling methods.
However, analyzing both the resulting maps and runtime configurations, two fundamental limitations of VSA-OGM emerge:
\begin{enumerate}
    \item Although the ToySim experiments suggest interpolation across sparse regions, performance degrades when data is highly imbalanced (e.g., dense hallways versus sparse rooms). This highlights sensitivity to non-uniform spatial sampling.
    \item As point counts increase by orders of magnitude, memory vectors become saturated, forcing us to increase the number of tiles per dimension ($\delta$) to maintain accuracy. Thus, vector dimensionality must scale with dataset size, or the number of tiles per dimension must be increased accordingly. This is a limitation we would like to address in future works by developing an information theoretic method to maintain consistent mapping performance while avoiding memory saturation.
\end{enumerate}

These limitations highlight open challenges within VSA-OGM, such as balancing dimensionality with information density. We will develop information-theoretic strategies for adaptive dimensionality allocation, ensuring that VSA-OGM preserves accuracy without saturating memory vectors.

\subsubsection{Intel Map - Multi Agent} \label{sec:intel-fusion}
{
\renewcommand{\arraystretch}{1.2}
\begin{table}[]
\centering
\caption{\textcolor{blue}{The performance of the individual VSA-OGM agents on the multi-agent Intel Map~\cite{Radish} dataset.}}
\begin{tabular}{c|cc|c}
\hline
\textbf{Quad.} & \textbf{AUC} & \textbf{NLL} & \textbf{Latency (s)} \\ \hline
$1$ & $0.95$ & $0.45$ & $0.04$ \\
$2$ & $0.96$ & $0.39$ & $0.04$ \\
$3$ & $0.94$ & $0.42$ & $0.04$ \\
$4$ & $0.93$ & $0.37$ & $0.04$ \\ \hline
Fusion & $0.95$ & $0.37$ & $0.03$ \\ \hline
\end{tabular}
\label{tab:intel-map-fusion-indv-stats}
\end{table}
}
{
\renewcommand{\arraystretch}{1.2}
\begin{table}[]
\begin{center}
\caption{The multi-map fusion capabilities of VSA-OGM versus Fast Bayesian Hilbert Maps~\cite{zhi2019continuous} based on AUC score \cite{BRADLEY19971145} and the learned model sizes. The reduction of model size within \cite{zhi2019continuous} is due to their approach assuming independence between \textcolor{blue}{cells}. (-) Values are not provided in the original manuscript.}
\begin{tabular}{c|ccc}
\hline
    \textbf{Metric}    & \textbf{VSA-OGM} & \textbf{Fast-BHM}~\cite{zhi2019continuous}  & \textbf{OHM}~\cite{doherty2016probabilistic} \\ \hline
\textbf{AUC}        & $0.95$  & $0.936$ & $0.92$ \\
\textbf{Model Size} & 16.4 MB & 0.04 MB & -     \\
\textbf{Covariance} & True    & False   & False     \\ \hline
\end{tabular}

\label{tab:intel-fusion-results}  
\end{center}
\end{table}
}
\begin{figure}
    \centering
    \includegraphics[width=0.75\linewidth]{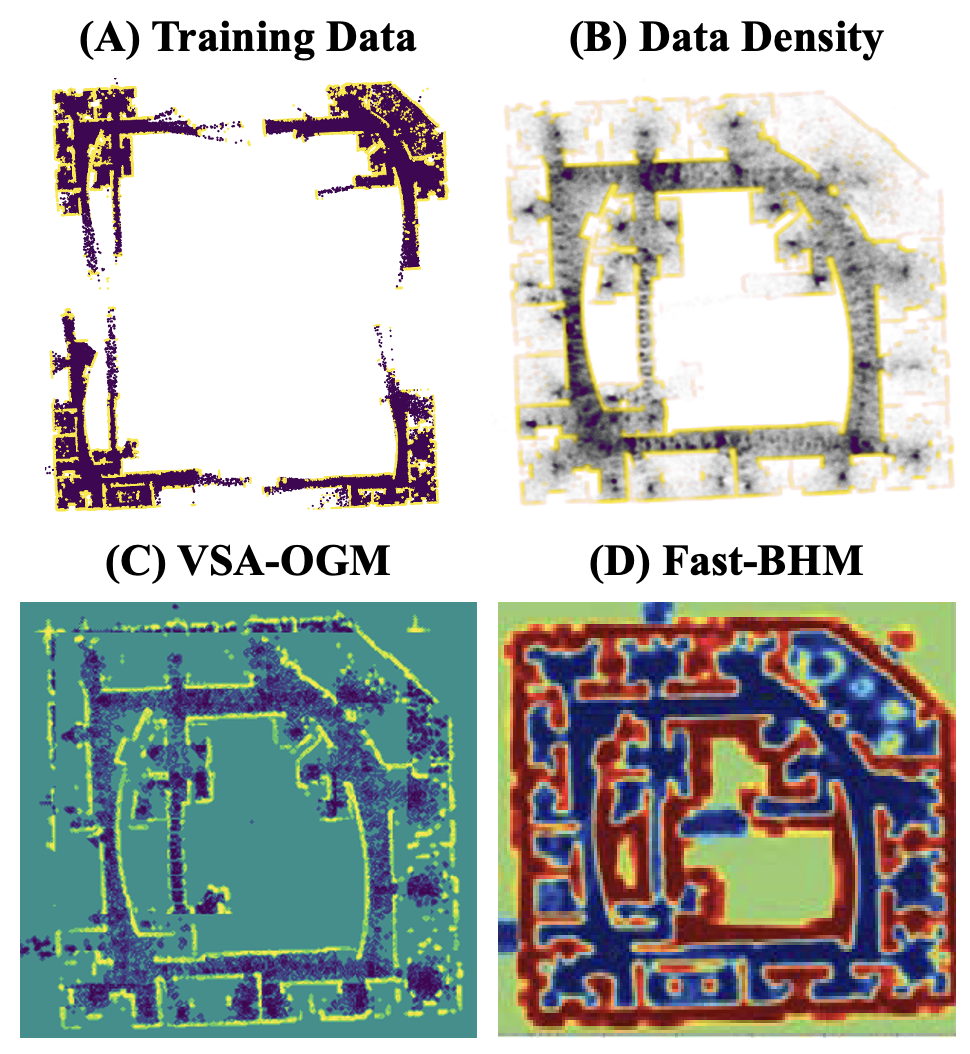}
    \caption{Multi-Agent fusion experiment: (A) The four quadrant occupancy grid maps with our method on the Intel Map dataset~\cite{Radish}. \textcolor{blue}{(B) The density of the training data. (C) The global combined OGM from VSA-OGM. (D) The final OGM from Fast-BHM~\cite{zhi2019continuous}. }}.
    \label{fig:intel-fusion-maps}
\end{figure}

Based on the multi-map fusion technique described in \cite{zhi2019continuous}, we split the Intel Map dataset~\cite{Radish} into $4$ quadrants across the $x-y$ plane. The splits of the global map are highlighted in Figure~\ref{fig:intel-fusion-maps}. \textcolor{blue}{A quantitative summary of the data is shown in Table~\ref{table:ma-data-summary}}.
Similar to Section~\ref{sec:sim-data-fusion}, all agents are equipped with identically configured hyperdimensional mapping systems. Therefore, each system is initialized with a vector dimensionality of \textcolor{blue}{$32000$}, a length scale of $0.2$ meters, disk radii of $1$, and \textcolor{blue}{$8$} quadrants per dimension. Individually, the final AUC values on the \textcolor{blue}{$20\%$} validation set were \textcolor{blue}{0.95, 0.96, 0.94, and 0.93.} Furthermore, the global fused AUC was \textcolor{blue}{0.95} compared to Fast-BHM~\cite{zhi2019continuous} with a final AUC of $0.94$. 
Our results also include an experiment performed in Fast-BHM~\cite{zhi2019continuous}, where they compare against another method known as Overlapping Hilbert Maps (OHM)~\cite{doherty2016probabilistic}. Compared to OHM, VSA-OGM improves the fused AUC by $0.03$.
Therefore \textcolor{blue}{VSA-OGM} achieves \textcolor{blue}{improved} performance to Fast-BHM \textcolor{blue}{and OHM} in multi-agent fusion. These values\textcolor{blue}{, as well as the per quadrant AUC and NLL values,} are summarized in Table~\ref{tab:intel-fusion-results} with the training data and OGMs visualized in Figure~\ref{fig:intel-fusion-maps}.

\subsubsection{EviLOG Dataset} \label{sec:evilog} Introduced in~\cite{evilog}, Evidential \textcolor{blue}{LiDAR} Occupancy Grid Mapping (EviLOG) uses a multi-layer convolutional neural network with $4.7$ million trainable parameters. \textcolor{blue}{Since they introduce a dataset and a network architecture, we refer to their dataset as EviLOG Dataset and their baseline network as EviLOG-BaseNet.} They introduce a fully-labeled synthetic \textcolor{blue}{OGM} dataset using a VLP32C LiDAR sensor mounted on the roof of a moving vehicle. \textcolor{blue}{The EviLOG dataset is separated into three sets for training, testing, and validation with each containing $10000$, $100$, and $1000$ point clouds, respectively. Each point cloud is bounded within a $56.32$ by $81.92$ meter rectangular region. Furthermore, each point cloud has a corresponding ground truth OGM with a \textcolor{blue}{cell} resolution of $0.16$ meters.}
{
\renewcommand{\arraystretch}{1.3}
\begin{table}[]
\centering
\label{tab:evilog-results}  
\caption{\textcolor{blue}{Comparing the training time and inference latency of VSA-OGM against EviLOG-BaseNet~\cite{evilog} and EviLOG-SmallNet.}}
\begin{tabular}{c|ccc}
\hline
\textbf{Model} & \textbf{\begin{tabular}[c]{@{}c@{}}Model\\Size\end{tabular}} & \textbf{\begin{tabular}[c]{@{}c@{}}Training\\Time\end{tabular}} & \textbf{\begin{tabular}[c]{@{}c@{}}Inference\\Latency\end{tabular}} \\ \hline
\textbf{EviLOG-BaseNet} & 4.7M & 30 hours & 290ms \\
\textbf{EviLOG-SmallNet} & 76K & 13 hours & 165ms \\
\textbf{VSA-OGM} & 64K & - & 55ms \\ \hline
\end{tabular}
\end{table}
}
\begin{figure}
    \centering
    \includegraphics[width=0.9\linewidth]{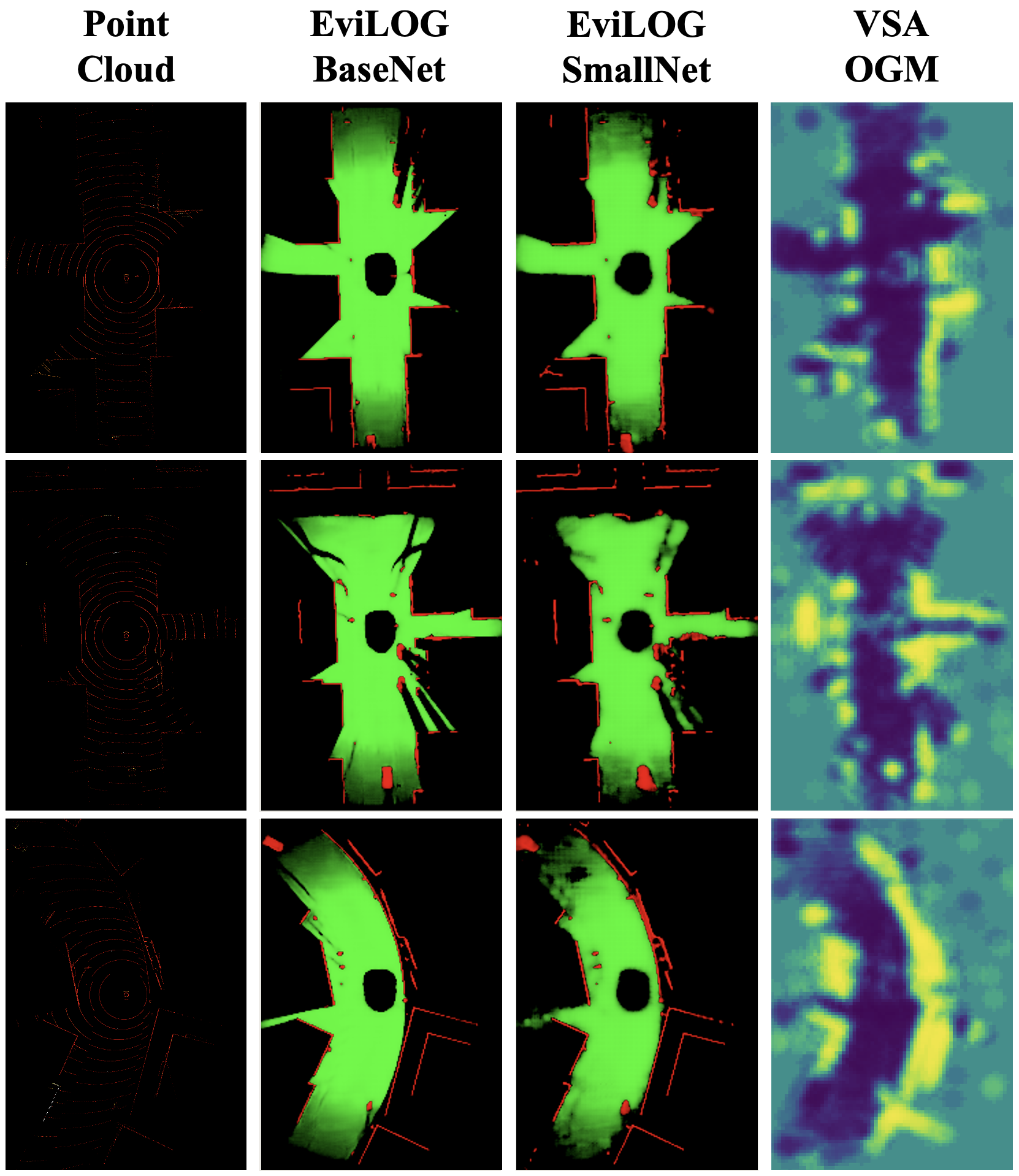}
    \caption{\textcolor{blue}{A comparison of VSA-OGM with EviLOG-BaseNet~\cite{evilog} and EviLOG-SmallNet. The input point clouds and decoded OGMs from each approach are highlighted.}}
    \label{fig:evilog-single}
\end{figure}

Training EviLOG-BaseNet for 100 epochs, with a batch size of 5, took approximately $30$ hours. This comes in contrast to \textcolor{blue}{VSA-OGM} where no pre-training is required and represents a limitation of neural methods. \textcolor{blue}{We created a small variant of EviLOG-BaseNet known as EviLOG-SmallNet. By reducing the number of features in the convolution layers by 8x, EviLOG-SmallNet has 76K model compared to VSA-OGM with 64K parameters. We performed the same training process with EviLOG-SmallNet and this process took approximately 13 hours.}

\textcolor{blue}{VSA-OGM} has a per frame inference time of \textcolor{blue}{55ms versus the EviLOG architectures at 165ms and 290ms for the small and original variants respectively}. Therefore without pretraining, VSA-OGM is approximately \textcolor{blue}{3x and 6x faster with a reduction in model parameters by 16\% and 99\%, respectively}. \textcolor{blue}{The configuration of VSA-OGM utilized in this experiment is shown in Table~\ref{tab:hyperparameters}.}

EviLOG-BaseNet calculates performance based on the KL-divergence across the entire domain, including areas with no \textcolor{blue}{labeled} points, so we are unable to compare via quantitative analysis with AUC and NLL (as in Section~\ref{sec:intel-global}). \textcolor{blue}{Therefore, we leverage qualitative analysis between the resulting OGMs.}

The EviLOG architectures begin with a PointPillars backbone~\cite{DBLP:journals/corr/abs-1812-05784}, before passing the latent representation \textcolor{blue}{through a }series of convolution layers.
\textcolor{blue}{This allows both EviLOG architectures to infer structure beyond directly observed points, leveraging priors learned during training to extrapolate into completely unobserved regions.}
This comes in contrast to \textcolor{blue}{VSA-OGM} which makes no prior assumptions about the observable space but uses the point labeling heuristic defined in the EviLOG paper~\cite{evilog}.
\textcolor{blue}{As shown in Figure~\ref{fig:evilog-single}, VSA-OGM produces accurate occupancy estimates in regions where observations exist but remain spatially sparse, however it struggles in areas that are completely unobserved.
This distinction suggests that VSA-OGM is well suited for partially observed settings, while the EviLOG architectures are more effective when strong priors are needed to extrapolate structure beyond the observed domain.}

% ----------------------------------
% Ablation Study
% ----------------------------------
\subsection{Ablation Study} \label{sec:ablation}

\textcolor{blue}{VSA-OGM} has multiple parameters that can be tuned for different datasets and performance requirements. In this section, we extensively explore the selection of these parameters.
\textcolor{blue}{We evaluate the contributions of the entropy-based decoding mechanism, isolating its effects to justify inclusion in VSA-OGM.}

\textcolor{blue}{We perform this ablation over three datasets: Ablation Dense, Ablation Sparse, and ToySim. As shown in Table}~\ref{table:data-summary}, \textcolor{blue}{these datasets vary by several orders of magnitude in density, which we define as the number of points per square meter. This variation is intentional, since dataset sparsity directly influences hyperparameter selection. Interpreting the results requires considering both the parameter settings themselves and their interaction with dataset density.}

While other dataset features such as class overlap, noise distribution, or temporal consistency may also affect performance, we focus here on the sparsity because it is the most quantifiable variation across the individual datasets.

Unless otherwise noted, all ablation experiments are performed over a range of hyperparameters defined in Table~\ref{tab-apx-decoding-params}. \textcolor{blue}{In the following subsections, we individually perform these ablations before concluding with a summary of major findings.}

{
\renewcommand{\arraystretch}{1.3}
\begin{table*}[]
\centering

\caption{\textcolor{blue}{The hyperparameter combinations evaluated for the ablation experiments. \textcolor{blue}{Ablation Dense and Ablation Sparse have a total of 288 parameter combinations. ToySim has a total of 384 parameter combinations.} The disk radii parameters are only used when the activation parameter is set to `Entropy`.}}
\begin{tabular}{c|ccc}
\hline
\textbf{Dataset}               & \textbf{\begin{tabular}[c]{@{}c@{}}Ablation\\ Sparse\end{tabular}} & \multicolumn{1}{c|}{\textbf{\begin{tabular}[c]{@{}c@{}}Ablation\\ Dense\end{tabular}}} & \textbf{ToySim}                \\ \hline

\textbf{Vector Dimensionality} & \multicolumn{3}{c}{8192, 16384, 32768}                       \\
\textbf{Vector Length Scale}   & \multicolumn{3}{c}{0.1, 0.2, 0.5, 1.0}                      \\ \hline
\textbf{Normalization}         & \multicolumn{3}{c}{Yes, No}                                                                                                                                                                   \\
\textbf{Non-Linearity}         & \multicolumn{3}{c}{Yes, No}                                                                                                                                                                   \\
\textbf{Activation}            & \multicolumn{3}{c}{ReLU, TanH, Sigmoid, Entropy}                                                                                                                                              \\ \hline
\textbf{Disk Radii 1}          & \multicolumn{2}{c|}{\multirow{2}{*}{1, 3, 5}}                                                                                                               & \multirow{2}{*}{3, 5, 7, 9, 11} \\
\textbf{Disk Radii 2}          & \multicolumn{2}{c|}{}                                                                                                                                       &                                 \\ \hline
\end{tabular}
\label{tab-apx-decoding-params}
\end{table*}
}

\subsubsection{Decoding Methodology} \label{section:decoding method}

\textcolor{blue}{We perform ablation experiments to highlight the performance benefits of applying linear, non-linear, and entropy-based post-processing steps to the baseline quasi-probability values ($\mathcal{P}$) from VSA-OGM. More information on the derivation of $\mathcal{P}$ can be found in Section~\ref{sec:decoding}. The entire process is broken into a toggle-able sequential processing paradigm defined as:
\begin{equation}
    \hat{\mathcal{P}}=f^{\delta_{ACT}}_{ACT}(f^{\delta_{NL}}_{NL}(f^{\delta_N}_{N}(\mathcal{P}))),
\end{equation}
where $\delta_N, \delta_{NL}, \delta_{ACT} \in \{0,1\}$ are binary toggles indicating whether normalization ($f_{N}$), non-linear functions ($f_{NL}$), and activations ($f_{ACT}$) are applied, respectively. Lastly, $\hat{\mathcal{P}}$ represents an intermediate representation of the true probability value ($Pr$) before the softmax function is applied.}

\textcolor{blue}{
For example, a function ($f_z$) with a toggle ($\delta_z$) over $\mathcal{P}$ is a defined as a piecewise function $f_z^{\delta_z}(\mathcal{P)}$:
\begin{equation}
    \hat{\mathcal{P}}=f_z^{\delta_z}(\mathcal{P})=
    \begin{cases}
        f_z(\mathcal{P}) & \text{if } \delta_z = 1 \\
        \mathcal{P} & \text{if } \delta_z = 0
    \end{cases}.
\end{equation}
}

\textcolor{blue}{Normalization ($f_N$) is performed by dividing the entire matrix by the maximum value: $\hat{\mathcal{P}} = \frac{\mathcal{P}}{\textrm{max}(\mathcal{P})}$. The nonlinear function ($f_{NL}$) is defined as a squaring operation, akin to the Born rule in quantum mechanics~\cite{landsman2009born}, where quantum states are connected with the probabilities of measured outcomes. Lastly, the activation function ($f_{ACT}$) is defined as one of the following: rectified linear unit (ReLU), hyperbolic tangent (TanH), sigmoid, and Shannon Entropy ($H$)~\cite{6773024}. More information on the implementation of Shannon entropy can be found in Section~\ref{sec:decoding}}.

\begin{figure}[h]
    \centering
    \includegraphics[width=\linewidth]{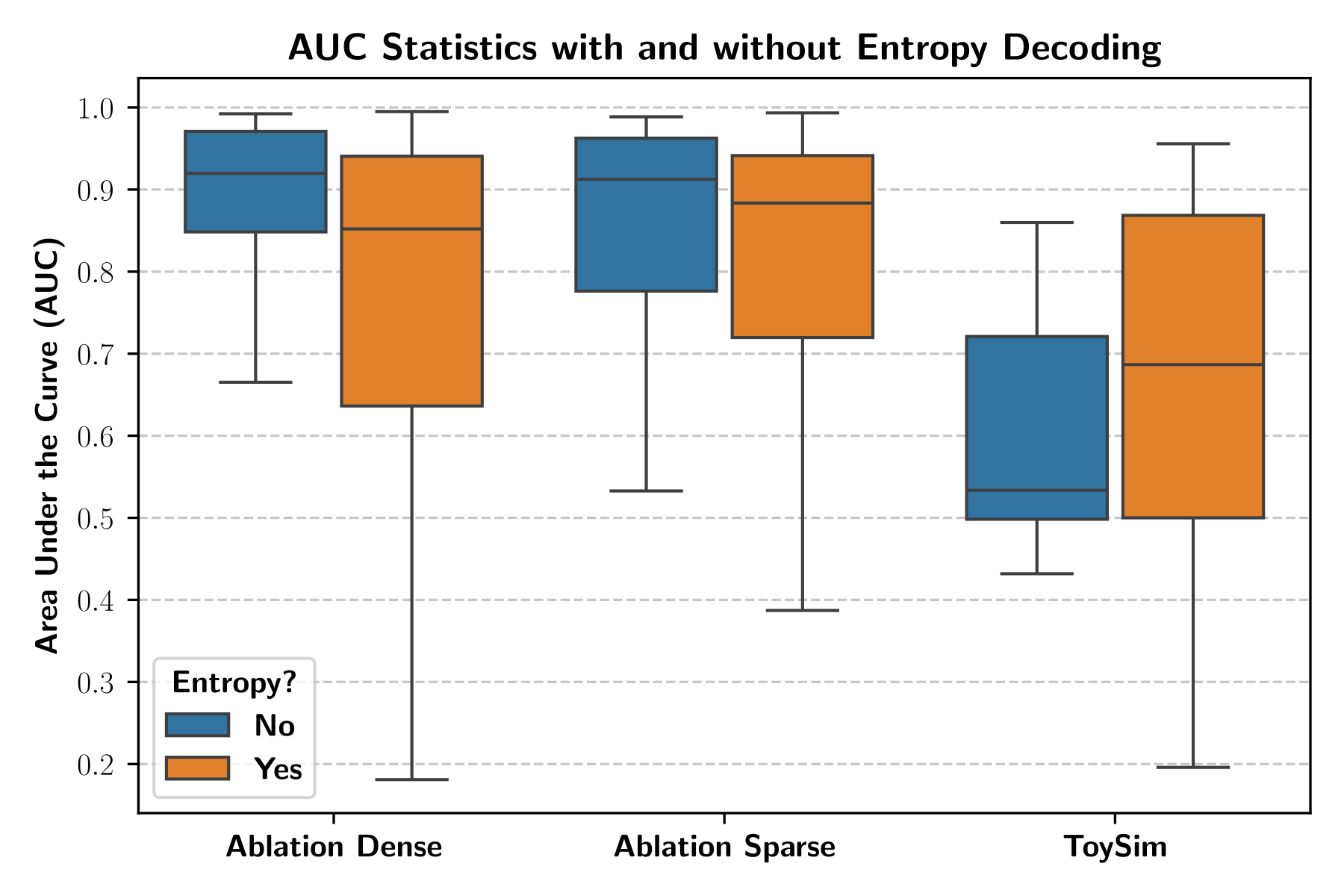}
    \caption{Comparing the Area under Receiver Operating Characteristic Curve with and without entropy-based decoding methodologies. \textcolor{blue}{Statistical analysis is performed over 288 parameter combinations with the Ablation datasets and 384 with ToySim. The individual parameters are defined in Table}~\ref{tab-apx-decoding-params}.}
    \label{fig:ablation-decoding-method-grouped-entropy-auc}
\end{figure}

\textcolor{blue}{These experiments are performed over a range of hyperparameters, shown in Table~\ref{tab-apx-decoding-params}, and three datasets \textcolor{blue}{with different sparsities}: Ablation Dense \textcolor{blue}{(densest)}, Ablation Sparse \textcolor{blue}{(intermediate)}, and ToySim \textcolor{blue}{(sparsest)}. More information on these datasets \textcolor{blue}{and the derivation of sparsity} is shown in Table~\ref{table:data-summary}.}
\textcolor{blue}{In the first segment, we evaluate the entropy versus non-entropy activations. In the second segment, we evaluate the performance characteristics of the normalization and non-linear functions. }

\textbf{Entropy and Non-Entropy Activations}
\textcolor{blue}{are evaluated across multiple hyperparameter combinations defined in Table~\ref{tab-apx-decoding-params}. These results are illustrated in Figure~\ref{fig:ablation-decoding-method-grouped-entropy-auc}.}
\textcolor{blue}{The influence of dataset sparsity is critical in interpreting these results.
In dense \textcolor{blue}{datasets}, entropy-based mechanisms exhibit a reduction in mean performance, reflecting both their heightened sensitivity to hyperparameters and the smoothing effect of entropy over local regions. In sparse datasets, however, this same smoothing becomes advantageous, leading to consistent improvements in mean performance across most parameter configurations, with only a subset producing degraded outcomes.
Together, these results demonstrate that the benefit of entropy-based decoding intensifies with dataset sparsity, as the smoothing effect provides greater value when observations are spatially sparse.
Section~\ref{section:ablation-entropy} further examines the specific entropy parameters that maximize performance in sparse datasets while mitigating penalties in dense datasets.}

\begin{figure}
    \centering
    \includegraphics[width=\linewidth]{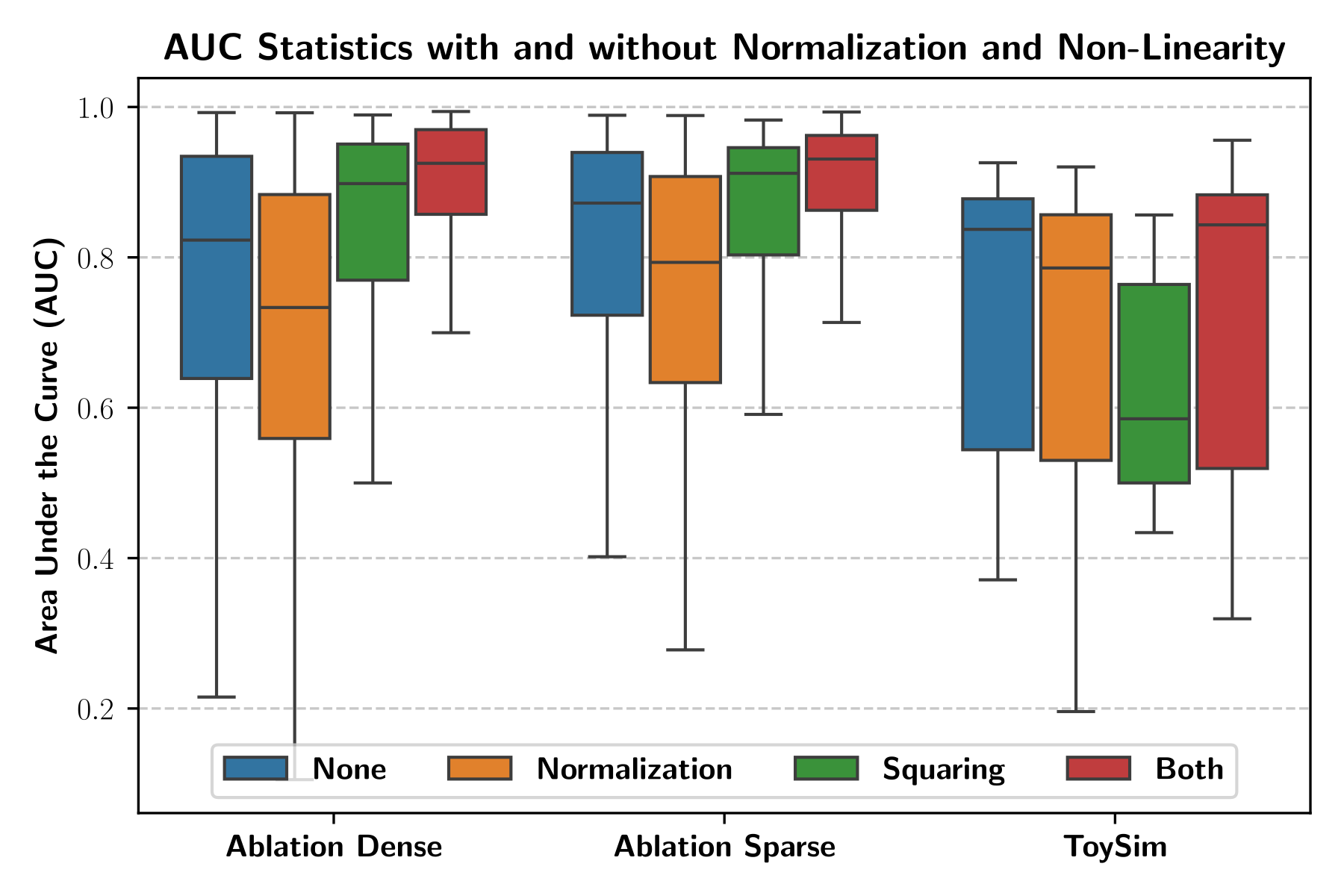}
    \caption{Comparing the Area under Receiver Operating Characteristic Curve with different parameters grouped based on the utilization of normalization and non-linearities. \textcolor{blue}{The statistics are calculated by performing multiple evaluations with the parameters defined in Table}~\ref{tab-apx-decoding-params}. \textcolor{blue}{The Ablation datasets have 288 combinations with ToySim having 384 combinations.}}
    \label{fig:ablation-decoding-method-grouped-norm-auc}
\end{figure}

\textbf{Normalization and Non-Linear Functions} \textcolor{blue}{are evaluated across a range of hyperparameters defined in Table~\ref{tab-apx-decoding-params}.
\textcolor{blue}{As shown in Figure}~\ref{fig:ablation-decoding-method-grouped-norm-auc}, the normalization function alone reduces the mean performance \textcolor{blue}{across all hyperparameter combinations and datasets in Table}~\ref{tab-apx-decoding-params}. This is caused by increased noise from the quasi-probability values being clustered around zero (see the quasi-probability values in Figure~\ref{fig:entropy-calculation}).  The non-linear squaring function improves the mean performance on the ablation datasets but reduces performance on ToySim. This is due to the sparsity of ToySim where the nonlinear squaring function reduces the effective kernel width and impedes interpolation to new environmental regions.
Notably, combining normalization with squaring consistently improves performance across all datasets, as the squaring function counteracts the noise amplification introduced by normalization.
The major takeaway from these results is that VSA-OGM configured with both normalization and the non-linear squaring function presents the highest mean performance across all three datasets.}

\subsubsection{Disk Filter Radii} \label{section:ablation-entropy}

\begin{figure}
    \centering
    \includegraphics[width=0.75\linewidth]{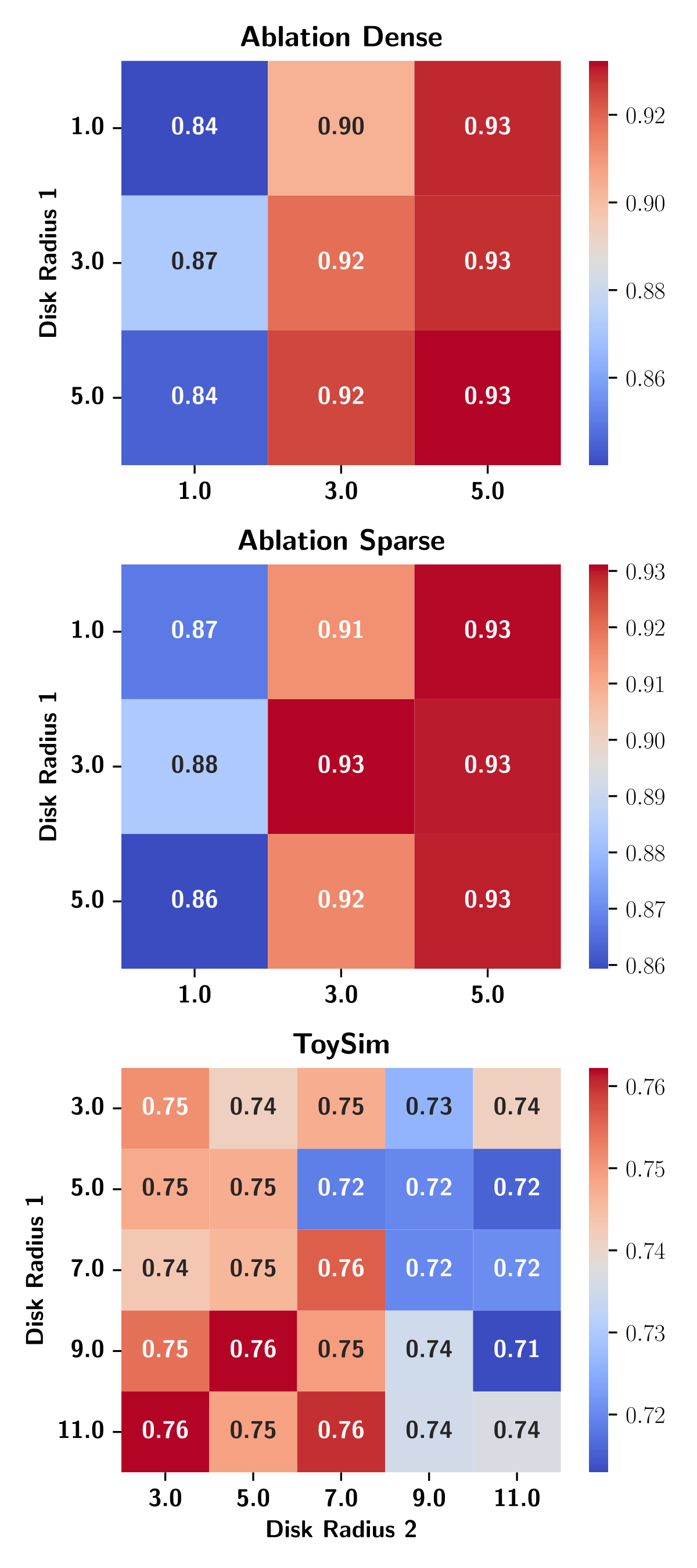}
    \caption{\textcolor{blue}{Evaluating the specification of the disk filter radii within the entropy decoding methods used in VSA-OGM across three datasets. The results represent the mean AUC scores across all parameter combinations in Table~\ref{tab-apx-decoding-params} with normalization and non-linear squaring.}}
    \label{fig:ablation-entropy-radii}
\end{figure}

\textcolor{blue}{The ablation study in Section~\ref{section:decoding method} highlighted that Shannon entropy decoding yielded the highest mean AUC on the ToySim dataset but reduced AUC on the Ablation Dense and Ablation Sparse datasets.
Entropy-based decoders introduce greater AUC variance across hyperparameter configurations, with some providing performance increases while others lead to performance degradation. In this ablation study, we systematically evaluate these additional parameters to identify a balanced configuration that preserves performance across each dataset.}

\textcolor{blue}{Entropy-based decoders introduce two additional parameters, which we denote as disk radius 1 (DR1) and disk radius 2 (DR2). DR1 and DR2 define the radius of the localized entropy mask filter (see Equation~\ref{eq:entropy-mask}) for the occupied and empty class, respectively. As shown in Figure~\ref{fig:ablation-entropy-radii} and detailed in Table~\ref{tab-apx-decoding-params}, we systematically adjust these parameters across the Ablation Dense, Ablation Sparse, and ToySim datasets. The heatmap values, in Figure~\ref{fig:ablation-entropy-radii}, report the mean AUC for each DR1 and DR2 combination, averaged over all hyperparameter configurations defined in Table~\ref{tab-apx-decoding-params}.}

\textcolor{blue}{Similar to Section~\ref{section:decoding method}, interpreting these results requires consideration of dataset \textcolor{blue}{sparsity}.
\textcolor{blue}{In this study, we denote two specific types of sparsity: dataset-level sparsity and intra-class sparsity. Dataset-level sparsity quantifies the number of points per square meter. On the other hand, intra-class sparsity refers the relative spatial densities of each class. For example, the occupied class is tightly clustered around environmental boundaries and the empty class defines broader traversable regions. Therefore, the occupied class has a low intra-class sparsity whereas the emtpy class has a high intra-class sparsity.}
More information on each dataset and their respective densities is shown in Table~\ref{table:data-summary}. As shown in Figure~\ref{fig:ablation-entropy-radii}, performance on Ablation Dense and Ablation Sparse presents sharper responses to parameter shifts, whereas the ToySim dataset is less sensitive. This suggests that spatially sparse datasets (i.e. ToySim) are, on average, less sensitive to different DR1 and DR2 values than spatially dense datasets (i.e. Ablation Dense and Ablation Sparse).}

\textcolor{blue}{
DR1 exhibits less performance variation across all parameter combinations and datasets than DR2. This arises because DR2 modulates the entropy decoder for the empty class (high intra-class sparsity), whereas DR1 modulates the occupied class (low intra-class sparsity). \textcolor{blue}{These results indicate that both dataset-level sparsity and inter-class sparsity influence the specification of DR1 and DR2}.
}

\textcolor{blue}{In summary, datasets with higher dataset-level densities are more sensitive to disk radii values because the high concentration of observations amplifies the effect of bundling noise, where successive bundling introduces additional noise. This problem is described further in Section~\ref{sec:decoding}. Within such datasets, however, the effect is distributed unevenly across classes. Datasets with high intra-class sparsity, such as the empty class in our datasets, exhibit greater variability because each observation presents more spatial information. Therefore, adjustments to DR2 alter how unobserved regions are interpolated from limited information, resulting in higher sensitivity. In contrast, the occupied class is buffered by redundancy where dense local sampling distributes the effect of radius variation across many redundant points.}

\textcolor{blue}{This resolves the apparent inconsistency that both denser datasets and sparser classes exhibit heightened sensitivity. Reduced dataset-level sparsity amplifies the overall magnitude of sensitivity, whereas intra-class sparsity dictates how evenly sensitivity is distributed.}

\subsubsection{Vector Length Scale} \label{section:ablation-vector-length-scale}

\begin{figure*}[t]
    \centering
    \includegraphics[width=0.75\linewidth]{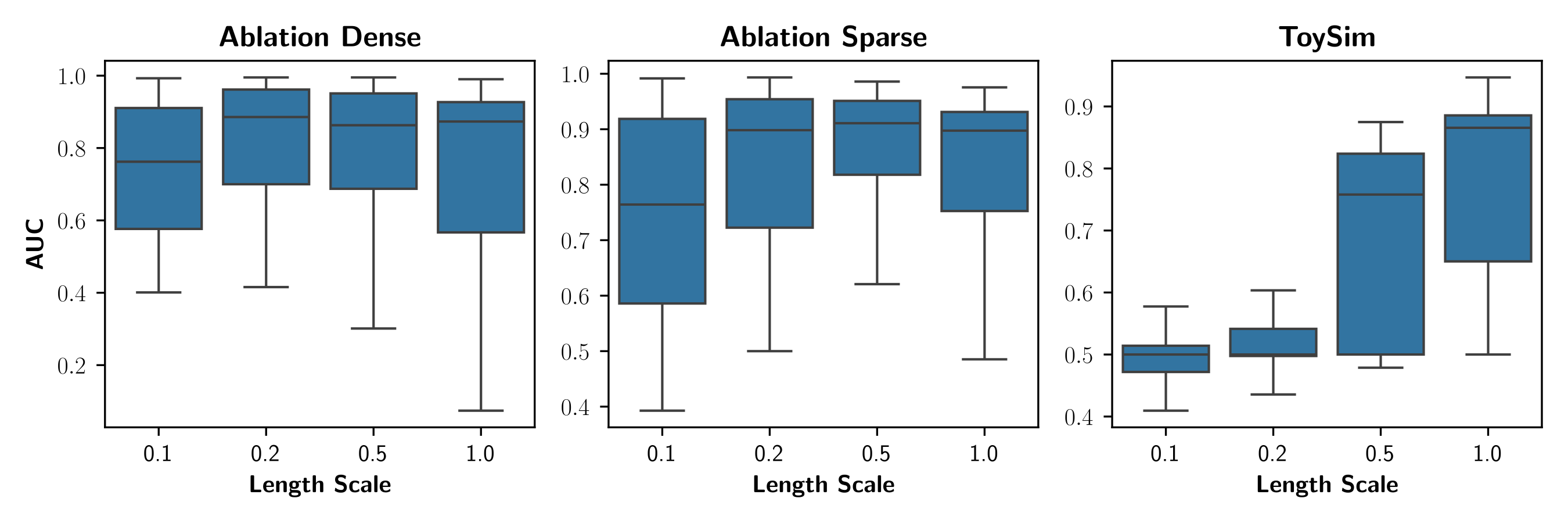}
    \caption{Evaluating multiple vector length scales compares to AUC scores across three datasets. The statistics are calculated across multiple parameter combinations defined in Table~\ref{tab-apx-decoding-params}. The ToySim dataset has 384 unique parameter combinations and the Ablation datasets have 288 combinations.}
    \label{fig:ablation_vector_length_scale_auc}
\end{figure*}

\textcolor{blue}{The vector length scale parameter \textcolor{blue}{adjusts the width the similarity kernel within VSA-OGM}. This parameter allows VSA-OGM's resolution per unit space to be adjusted based on \textcolor{blue}{dataset sparsity}.
As shown in Figure~\ref{fig:ablation_vector_length_scale_auc}, we evaluate \textcolor{blue}{the} vector length scale across three datasets: Ablation Dense, Ablation Sparse, and ToySim. More information on these datasets is shown in Table~\ref{table:data-summary}.}
\textcolor{blue}{As discussed in Section~\ref{section:ablation-entropy} and Section~\ref{section:ablation-vector-length-scale}, understanding the sparsity of each dataset is critical to interpret the results presented in Figure~\ref{fig:ablation_vector_length_scale_auc}.
}

\textcolor{blue}{Across the range of length scale values, the mean AUC of the Ablation Dense \textcolor{blue}{(densest)} and Ablation Sparse \textcolor{blue}{(intermediate density)} datasets remains relatively stable. On the contrary, the length scale parameter has a drastic impact on the mean AUC with the ToySim dataset (sparsest), where increasing the length scale parameter converts VSA-OGM from a classifier with performance no better than a random classifier (AUC = 0.50) to a high quality classifier with AUC scores greater than 0.95. The differing response to vector length scale across these datasets is caused by the \textcolor{blue}{variable sparsity within each dataset}. As shown in Table~\ref{table:data-summary}, ToySim has less training points than both the Ablation Dense and Ablation Sparse datasets while increasing the area of the environment by approximately two orders of magnitude. This highlights that the length scale parameter should be specified as a function of the overall environment size and data density, where larger environments with sparser data should have an increased length scale compared to smaller environments with denser data having a smaller length scale. Intuitively, this makes sense because increasing the vector length scale allows the quasi-kernel properties of VSA-OGM to expand across larger regions and make higher quality inferences further away from the training data, at the expense of localized performance.}

% Discuss the effects of the number of quadrants
\subsubsection{Environment Tiling} \label{section:ablation-tiling} While \textcolor{blue}{VSA-OGM} could be realized in a single, global memory vector, this would be impractical for real-time distributed systems with power, computational, and memory constraints. \textcolor{blue}{These issues arise as a result of increasing vector dimensionality ($d$) to accurately model larger and more intricate environments. Moreover, increasing $d$ raises the computational complexity of generating OGMs. In this section, we evaluate the effectiveness of VSA-OGM's environmental tiling mechanism to alleviate additional computational complexity by discretizing the environment into smaller tiles with each having a reduced $d$.}

{
\renewcommand{\arraystretch}{1.3}
\begin{table}[]
\centering
\caption{Evaluating the performance characteristics of VSA-OGM with the same global dimensionality across varying numbers of tiles per dimension $\delta$ on the ToySim dataset. The AUC and Latency values represent the mean of 3 separate runs.}
\begin{tabular}{c|cc|cc}
\hline
\textbf{Global $d$} & \textbf{Tile $d$} & $\delta$ & \textbf{AUC} & \textbf{Latency} \\ \hline
\multirow{6}{*}{$32000$} & 320   & 10 & 0.93 & 19.7ms \\
                         & 500   & 8  & 0.90 & 22.5ms \\
                         & 888   & 6  & 0.91 & 21.2ms \\
                         & 2000  & 4  & 0.94 & 21.2ms \\
                         & 8000  & 2  & 0.91 & 25.8ms \\
                         & 32000 & 1  & 0.92 & 66.8ms \\ \hline
\end{tabular}
\label{table:tiling-ablation}
\end{table}
}

\textcolor{blue}{As shown in Table~\ref{table:tiling-ablation}, we evaluate the performance and computational characteristics of VSA-OGM's environmental tiling mechanism on the ToySim dataset. More information on this dataset can be found in Table~\ref{table:data-summary}. Since environmental discretization is parameterized by $\delta$ tiles per axis, a 2D environment will have $\delta^2$ tiles.}

\textcolor{blue}{To \textcolor{blue}{ensure} a fair comparison between multiple parameter configurations, we define the global dimensionality (\textit{Global $d$}) and the tile dimensionality (\textit{Tile $d$}). \textit{Global} $d$ specifies the cumulative $d$ across all tiles, whereas \textit{Tile $d$} specifies the individual $d$ of each tile. For example, a 2D environment with 2 tiles  per axis (i.e. $\delta=2$) would have 4 tiles in total. If each tile had a \textit{Tile $d$} of $128$, the \textit{Global $d$} would be $128 * 4=512$.}

\textcolor{blue}{As shown in Table~\ref{table:tiling-ablation}, the \textit{Global $d$} for the ToySim dataset is configured at 32000.}

\textcolor{blue}{We evaluate multiple values of $\delta$, as this parameter directly determines the resulting \textit{Tile $d$} and thereby ensures that the \textit{Global $d$} constraint is satisfied.
We evaluate a range of $\delta$ values from 1 to 10 which reduces the \textit{Tile $d$} from 32000 to 320. All AUC and latency \textcolor{blue}{values in} Table~\ref{table:tiling-ablation} represent the mean of three individual runs.}

\textcolor{blue}{The major finding from this ablation is that increasing $\delta$ to 10 yields a $70.5\%$ reduction in mean latency while maintaining comparable accuracy. This demonstrates that reducing \textit{Tile $d$} effectively lowers the computational complexity of VSA-OGM without compromising performance. Intuitively, smaller \textit{Tile $d$} values reduce the number of multiply–accumulate operations (MACs) in proportion to the reduction in $d$. As shown in Table~\ref{table:tiling-ablation}, however, the relationship between $\delta$ and latency is not strictly linear. The largest improvement occurs when $\delta$ is increased from 1 to 2, resulting in a $61\%$ reduction in latency. By contrast, increasing $\delta$ to 10 provides only a modest improvement, reaching a $70.5\%$ reduction. \textcolor{blue}{These results indicate the presence of an upper bound}, beyond which further increases in $\delta$ yield diminishing latency reductions.}
\section{Discussion \& Future Works}

In this paper, we present an approach to occupancy grid mapping (OGM) drawing inspiration from recent advances in cognitive science~\cite{plate2003holographic, dumont2025symbols} and computational neuroscience~\cite{eliasmith2013build}. Our methodology is grounded in the evolving field of vector symbolic architectures (VSA)~\cite{Kleyko_2022}. VSAs have been conceptualized as a computationally viable abstraction for modeling biological systems, including memory, cognition, and motor control~\cite{eliasmith2013build}.

This research builds upon the work of \cite{eliasmith2003neural} which introduced Spatial Semantic Pointers (SSP), an algebraic framework founded on real-valued hyperdimensional vectors and circular convolution. Previous studies demonstrate the efficacy of SSP-based VSA systems serving as kernel density estimators in hyperdimensional space~\cite{voelker2020short}. Expanding upon these prior works, we describe an end-to-end \textcolor{blue}{probabilistic} OGM framework founded on SSPs, \textcolor{blue}{known as} VSA-OGM.

Existing \textcolor{blue}{OGM} approaches encounter two prominent challenges. Firstly, traditional methods offer high accuracy but struggle with significant time complexity due to the learning and parameterization of fully specified statistical distributions. Secondly, neural methods demonstrate high accuracy in extrapolating beyond observed domains with reduced latency but have limited application in mission and safety critical applications due to operating stochasticity in unseen domains. \textcolor{blue}{Moreover, neural methods are constrained by domain-specific pretraining which limits their application in novel domains without repeating the pre-training process.}

In response to these challenges, \textcolor{blue}{VSA-OGM} pioneers a novel direction in OGM systems. We refer to this new direction as neuro-symbolic methods since they draw inspiration from \textcolor{blue}{traditional} and neural \textcolor{blue}{methods}. Neuro-symbolic methods seek to leverage the determinism and high accuracy of \textcolor{blue}{traditional methods} while harnessing the enhanced computational efficiency of neural methods. By amalgamating the strengths of both domains, \textcolor{blue}{neuro-symbolic} methods offer a promising solution to address the computational complexity and stochasticity issues prevalent in current mapping techniques.

We showcase the efficacy of VSA-OGM as a \textcolor{blue}{neuro-symbolic} method through comprehensive evaluation, including an ablation study, quantitative, and qualitative comparisons with state-of-the-art \textcolor{blue}{traditional} \cite{senanayake2017bayesian, zhi2019continuous, duong2022autonomoussbkm, doherty2016probabilistic, Ghaffari_Jadidi2018-vg, elfes1989using} and neural methods \cite{evilog}. In our ablation study, we assess the influence \textcolor{blue}{of our entropy-based decoding mechanism, its associated parameters, the environment discretization mechanism, and the vector length scale.}
Our findings illuminate the impact of these parameters on the accuracy and stability of system performance. Our results and analysis underscore the robustness and adaptability of \textcolor{blue}{VSA-OGM}, positioning it as a promising alternative that bridges the strengths of both \textcolor{blue}{traditional} and neural methods in addressing complex spatial mapping tasks.

Compared to the \textcolor{blue}{traditional} methods Bayesian Hilbert Maps (BHM) \cite{senanayake2017bayesian} and Fast Bayesian Hilbert Maps (Fast-BHM) \cite{zhi2019continuous}, \textcolor{blue}{VSA-OGM} achieves similar levels of mapping accuracy while reducing inference latency by approximately \textcolor{blue}{45}x and \textcolor{blue}{5.5}x, respectively. Furthermore, \textcolor{blue}{VSA-OGM} facilitates multi-map fusion among multiple agents. Our results demonstrate minimal information loss during the fusion process and performance comparable to that of Fast-BHM. This capability highlights the versatility and scalability of \textcolor{blue}{VSA-OGM} in collaborative, multi-agent mapping scenarios.

In comparison to the neural method presented in EviLOG~\cite{evilog}, VSA-OGM demonstrates the capability to accurately model and interpolate LiDAR points sampled along major road sections, and around moving vehicles without the need for pretraining sessions. This highlights the efficiency and effectiveness of \textcolor{blue}{VSA-OGM} in capturing complex environmental dynamics. However, it's important to note a key distinction between VSA-OGM and EviLOG: while \textcolor{blue}{VSA-OGM} excels in accurately representing data within the observed domain, it encounters challenges in extrapolating outside this domain, particularly in areas with extremely limited information. This limitation arises from \textcolor{blue}{VSA-OGM's} fundamental principle of not assuming any specific properties about the operating environment. It is worth noting that EviLOG may be more susceptible to domain adaptation issues, as it is finely tuned to extrapolate domain-specific information.

Moving forward, our aim is to expand VSA-OGM to accommodate semantic 3D applications~\cite{jiang2020rellis3d}. Furthermore, we intend to explore strategies for mitigating the performance limitations associated with VSA-OGM's lack of assumptions about the environment \textcolor{blue}{and efficiently managing the limited memory vector capacity as data density increases}.
\section*{Author Contributions}

\noindent
SS developed the idea, performed design of experiments, analyzed the results, and wrote the manuscript.  AC and DG helped with analysis and reviewing the manuscript. MP contributed to the idea, analysis, writing the manuscript, and  supported/funded the project.

\section*{Data Availability}

\noindent
The datasets that where used in each experiment are available here: \href{https://github.com/Parsa-Research-Laboratory/VSA-OGM}{https://github.com/Parsa-Research-Laboratory/VSA-OGM}.

\section*{Code Availability}

\noindent
The code for this study is available here: \href{https://github.com/Parsa-Research-Laboratory/VSA-OGM}{https://github.com/Parsa-Research-Laboratory/VSA-OGM}.
\section*{Acknowledgment}

\noindent
\textit{We acknowledge the technical and financial support of
the Automotive Research Center (ARC) in accordance with Cooperative Agreement W56HZV-19-2-0001 U.S. Army DEVCOM Ground Vehicle Systems Center (GVSC) Warren, MI. This project was also supported by resources provided by the Office of Research Computing at George Mason University (URL: https://orc.gmu.edu) and funded in part by grants from the National Science Foundation (Award Number 2018631).}
\bibliographystyle{IEEEtran}
\bibliography{references}

\end{document}